\documentclass[10pt]{amsart} 
\usepackage{latexsym} 
\usepackage{amssymb} 
\usepackage{amscd}
\usepackage{epsf} 
\usepackage{graphicx}
\usepackage{verbatim}
\usepackage{longtable}
\usepackage{setspace}

\newtheorem{Thm}{Theorem}[section]

\newtheorem{Corollary}[Thm]{Corollary} 
\newtheorem{Proposition}[Thm]{Proposition}

\theoremstyle{definition} 
\newtheorem{Definition}[Thm]{Definition} 
\theoremstyle{definition} 
\newtheorem{Example}[Thm]{Example} 
 
\newtheorem{Remark}[Thm]{Remark} 

\numberwithin{equation}{section}

\newcommand{\R}{{\mathbb R}}

\newcommand{\flushpar}{\par \noindent}

\newcommand{\vol}{{\rm vol}\,}

\newcommand{\bdyM}{\partial M}

\def \bn {\mathbf {n}} 
\def \bs {\mathbf {s}}
 
\def \bu {\mathbf {u}} 
\def \bv {\mathbf {v}}

\def \bw {\mathbf {w}} 
\def \b1 {\mathbf {1}}

\def \bgW {\boldsymbol \Omega}

\def \cB {\mathcal{B}}

\def \cI {\mathcal{I}}

\def \cN {\mathcal{N}}

\def \cR {\mathcal{R}}

\def \ga {\alpha}

\def \gk {\kappa} 
\def \gl {\lambda}

\def \gG {\Gamma} 
 
\def \gL {\Lambda} 
 
\def \gW {\Omega}

\begin{document} 

\title[Shape and Positional Geometry] {Shape and Positional Geometry of 
Multi-Object Configurations} 
\author[James Damon and Ellen Gasparovic ]{James Damon$^1$ and Ellen 
Gasparovic$^2$} 
\thanks{(1) Partially supported by the Simons Foundation grant 230298, 
the National Science Foundation grant DMS-1105470 and DARPA grant 
HR0011-09-1-0055. (2) This paper contains work from this author's Ph. D. 
dissertation at Univ. of North Carolina.} 
\address{Dept. of Mathematics \\ 
University of North Carolina \\ 
Chapel Hill, NC 27599-3250 }
\email{jndamon@math.unc.edu}
\address{Dept. of Mathematics \\
 Union College \\
Schenectady, NY 12308
}
\email{gasparoe@union.edu}

\begin{abstract}
In \cite{DG1}, we introduced a method for modeling a configuration of 
objects in 2D and 3D images using a mathematical ``medial/skeletal 
linking structure.''  In this paper, we show how these structures 
allow us to capture positional properties of a multi-object configuration 
in addition to the shape properties of the individual objects.  
In particular, we introduce numerical invariants 
for positional properties which measure the closeness of neighboring 
objects, including identifying the parts of the objects which are close, 
and the ``relative significance'' of objects compared with the 
other objects in the configuration.  Using these numerical measures, we 
introduce a hierarchical ordering and relations between the individual 
objects, and quantitative criteria for identifying subconfigurations.  In 
addition, the invariants provide a ``proximity matrix'' which 
yields a unique set of weightings measuring overall proximity of objects 
in the configuration.    
Furthermore, we show that these invariants, which are volumetrically 
defined and involve external regions, may be computed via integral 
formulas in terms of ``skeletal linking integrals'' defined on the 
internal skeletal structures of the objects.  
\end{abstract}

\keywords{Blum medial axis, skeletal structures, spherical axis, Whitney 
stratified sets, medial and skeletal linking structures, generic linking 
properties, model configurations, radial flow, linking flow,  measures of 
closeness, measures of significance, proximity matrix, proximity weights, 
tiered linking graph} 
\subjclass{Primary: 53A07, 58A35, Secondary: 68U05}

\maketitle

\section{Introduction}\label{S:sec0}

In many 2D and 3D images, such as medical images, there appears a 
configuration of objects, and the analysis of objects in the image benefits 
from modeling the interplay of the different objects' shapes and their 
relative positions.  First steps for such an approach for medical images 
was begun by the MIDAG group at UNC led by Pizer, see e.g. \cite{JSM}, 
\cite{LPJ}, \cite{JPR}, \cite{GSJ}, and \cite{CP}.  These results use a 
modification of the classical Blum medial axis to model the individual 
regions together with user chosen, somewhat ad hoc, approaches to 
relating neighboring objects.  Results for the Blum medial axis of an
individual region, introduced by Blum-Nagel \cite{BN}, have concerned its 
generic structure using a number of different approaches (see, e.g., Mather 
\cite{M}, Yomdin \cite{Y}, Kimia et al \cite{KTZ}, Giblin and Kimia 
\cite{Gb}, \cite{GK}), and its computation (see, e.g., for \lq\lq grassfire 
flow\rq\rq\, Siddiqi et al. \cite{SBTZ}, the surveys by Pizer et al. \cite{P} and 
\cite{PS} including Voronoi methods, and for b-splines, Musuvathy et al.
\cite{MCD}).  The modification uses methods from \lq\lq skeletal 
structure\rq\rq models for objects as single regions with smooth 
boundaries (for 2D and 3D see \cite{D3} or \cite{D5} and more generally 
\cite{D1}, \cite{D2}).  These results add considerable flexibility and 
stability to the classical Blum medial axis.
  \par
In \cite{DG1} we introduced for configurations of objects in $\R^2$ or 
$\R^3$ medial/skeletal linking structures which capture both the 
shapes of the individual objects and their relative positions in the 
configuration.  In this paper we develop an approach to the \lq\lq 
positional geometry\rq\rq\, of a configuration using mathematical tools 
defined in terms of the linking structure, which build upon the methods 
already developed for skeletal structures for single regions.  Moreover, we 
will see that certain constructions and operators defined for skeletal 
structures and used for determining the geometry of individual objects 
can be extended to give simultaneously the positional geometric 
properties of the entire configuration.  As such this provides a natural 
progression from individual objects to configurations of objects.  \par
\begin{figure}[!t]
\begin{center} 
\includegraphics[width=2.5cm]{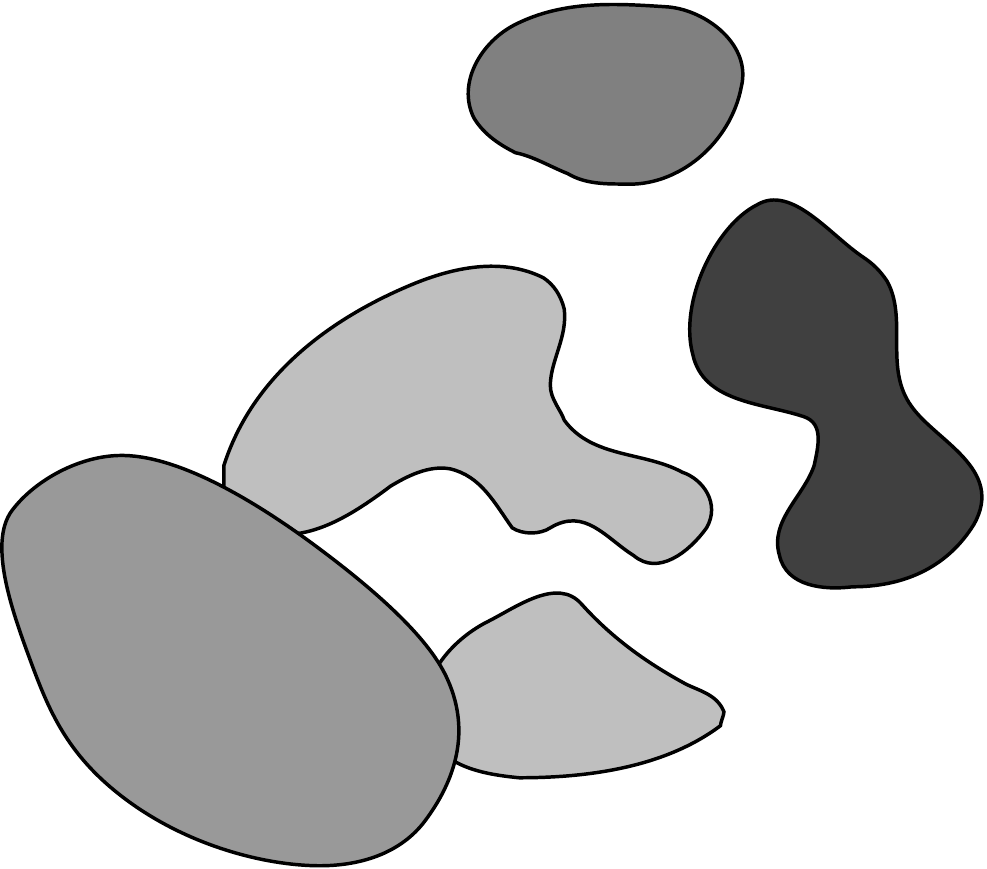}
\hspace*{1.5cm}
\includegraphics[width=2.5cm]{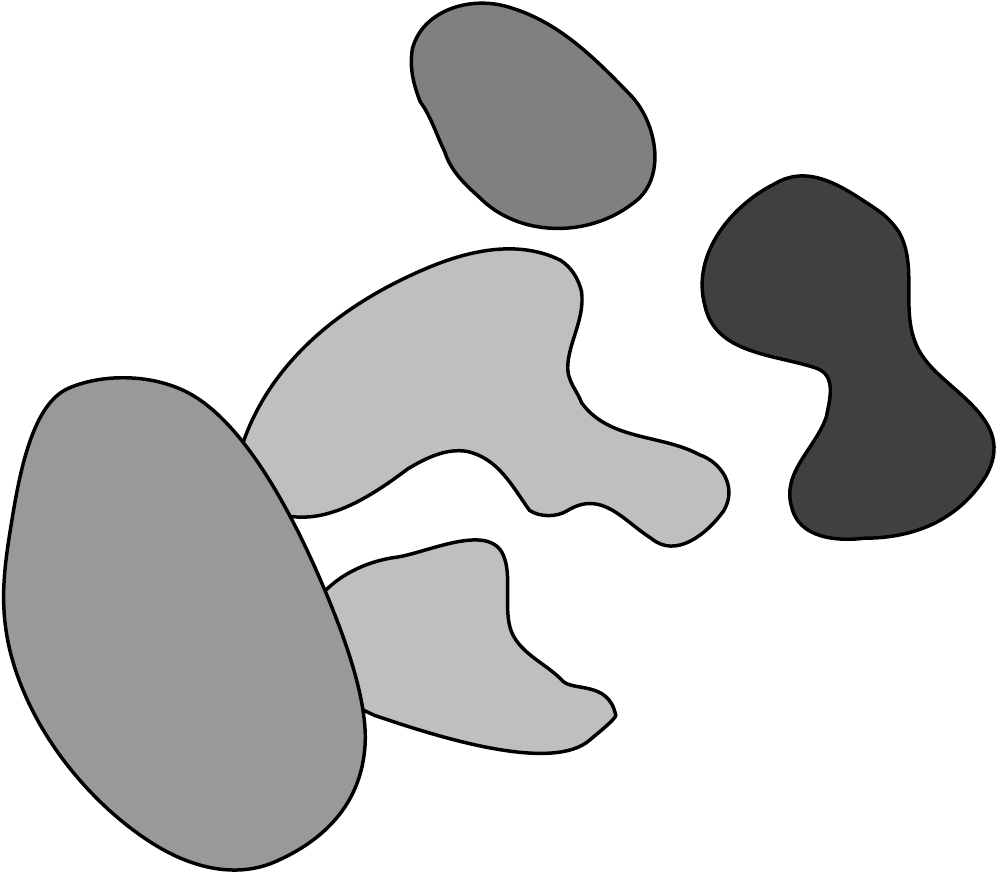}
\end{center}
a) \hspace{1.5in} b) \hspace {1.5in}
\caption{\label{fig.II1} Images exhibiting configurations of five objects.  
A basic problem is to determine the differences between the 
configurations that are due to changes in the shapes of the objects versus 
those due to position changes.  Furthermore, one would want to find 
numerical invariants which measure these differences.}  
\end{figure}
\par
Given a collection of configurations, we may ask what are the 
statistically meaningful shared geometric properties of the collection of 
configurations, and how the geometric properties of a particular 
configuration differ from those for the collection.  To provide quantitative 
measures for these properties, we will directly associate geometric 
invariants to a configuration.  Such invariants may be globally 
defined depending on the entire configuration or locally defined invariants 
depending on local subconfigurations associated to each object.
\par
For example, if we view the union of the objects as a topological space, 
then we can measure the Gromov-Hausdorff distance between two such 
configurations.  We may also use the geodesic distance  between the two 
configurations measured in a group of global diffeomorphisms mapping one 
configuration to another.  Such invariants give a single numerical global 
measure of differences between two configurations.  Instead, we will use 
skeletal linking structures associated to the configurations to directly 
associate both global and local geometric invariants which can be used to 
measure the differences between a number of different features of 
configurations in a variety of different ways.  \par
\begin{figure}[!t]
\centering
\includegraphics[width=2.5cm]{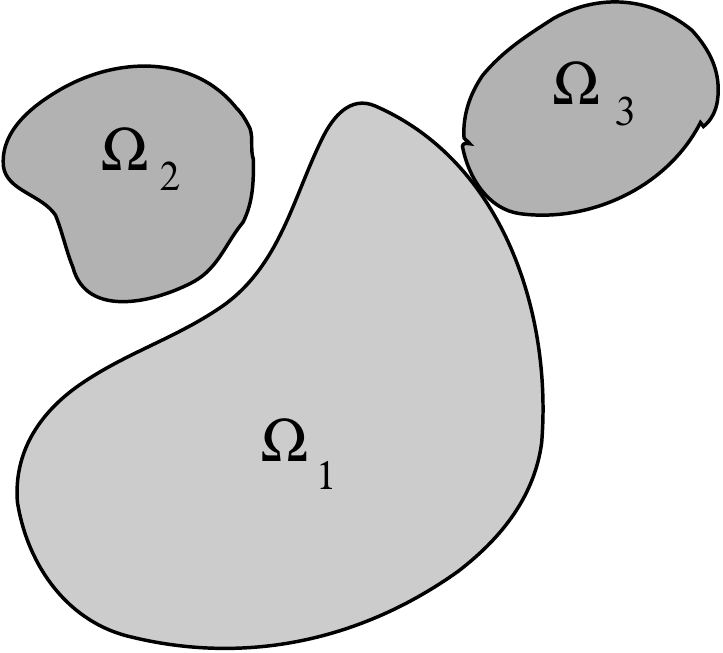}
\caption{\label{fig.II2} Measuring closeness of objects in a configuration.  
Although $\gW_3$ touches $\gW_1$, only a small portion of $\gW_3$ is 
close to $\gW_1$.  In contrast, $\gW_2$ does not touch $\gW_1$, but it 
remains close over a larger region.  Measuring closeness requires 
including both of these contributions, both locally and globally.}
\end{figure}

In introducing these invariants, we will be guided by several key 
considerations.  The first involves distinguishing between the differences 
in the shapes of individual objects versus their positional differences and 
how each of these contributes to the differences in the configurations, as 
illustrated in Figure~\ref{fig.II1}.  A first question for objects that do not 
touch is when they should be considered \lq \lq neighbors\rq \rq and what 
should be the criterion?  Second, in measuring the relative positions of 
neighboring objects, more than just the minimum distance between their 
boundaries is required; we also wish to measure how much of the regions 
are close, see, e.g., Figure~\ref{fig.II2}.  A goal is then to define numerical 
measures of {\it closeness of objects} which takes into account both 
aspects.  

\par
\begin{figure}[!t]
\begin{center} 
\includegraphics[width=2.5cm]{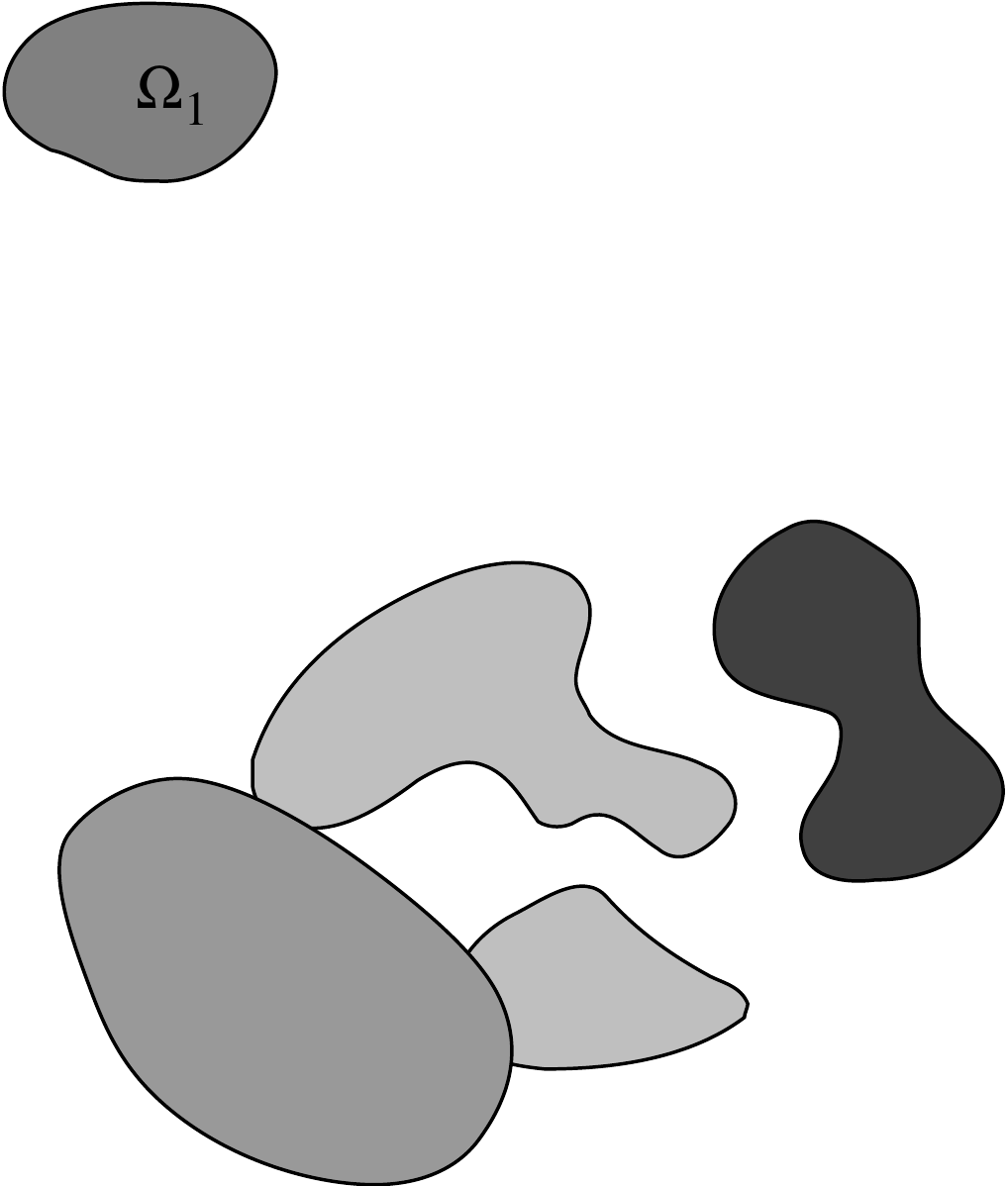} 
\hspace*{1.5cm}
\includegraphics[width=2.5cm]{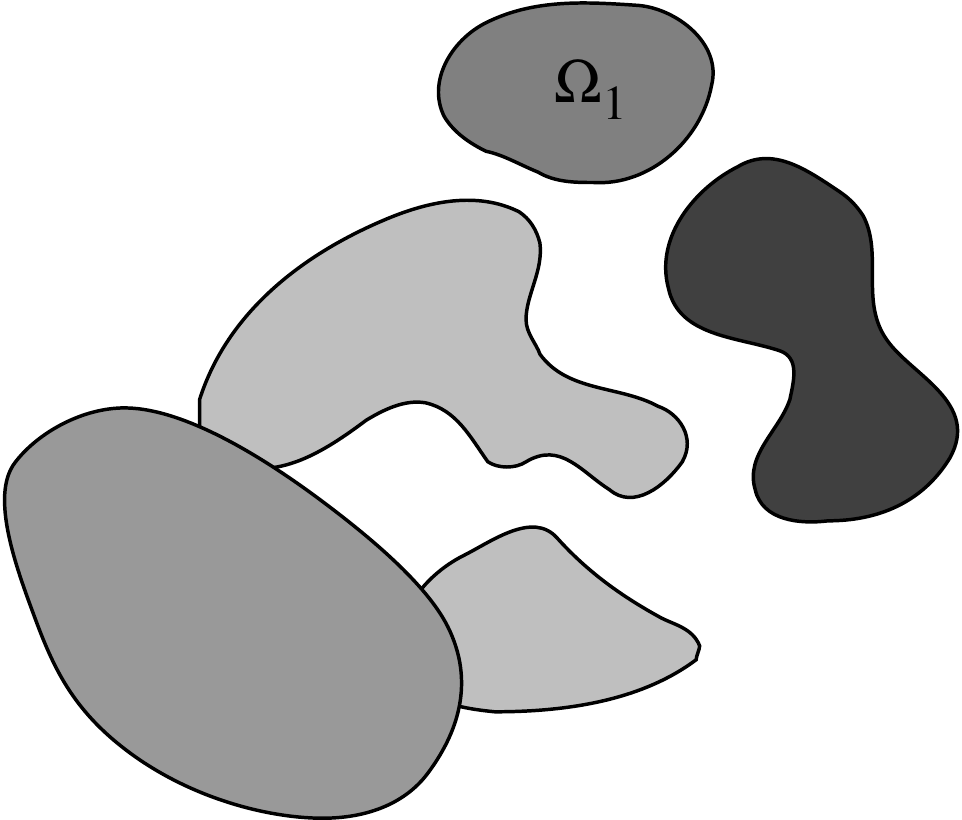}  
\end{center}  
a) \hspace{1.5in} b) \hspace {1.5in}

\caption{ \label{fig.II3} Images exhibiting configurations of five objects.  
In a), $\gW_1$ is a greater distance from the remaining objects, 
and hence is less significant when modeling positions within the 
configuration.  In b), the closeness of object $\gW_1$ to the other objects 
makes it more significant for modeling the positions within the 
configuration.}  
\end{figure}
\begin{figure}[!t]
\begin{center} 
\includegraphics[width=2.5cm]{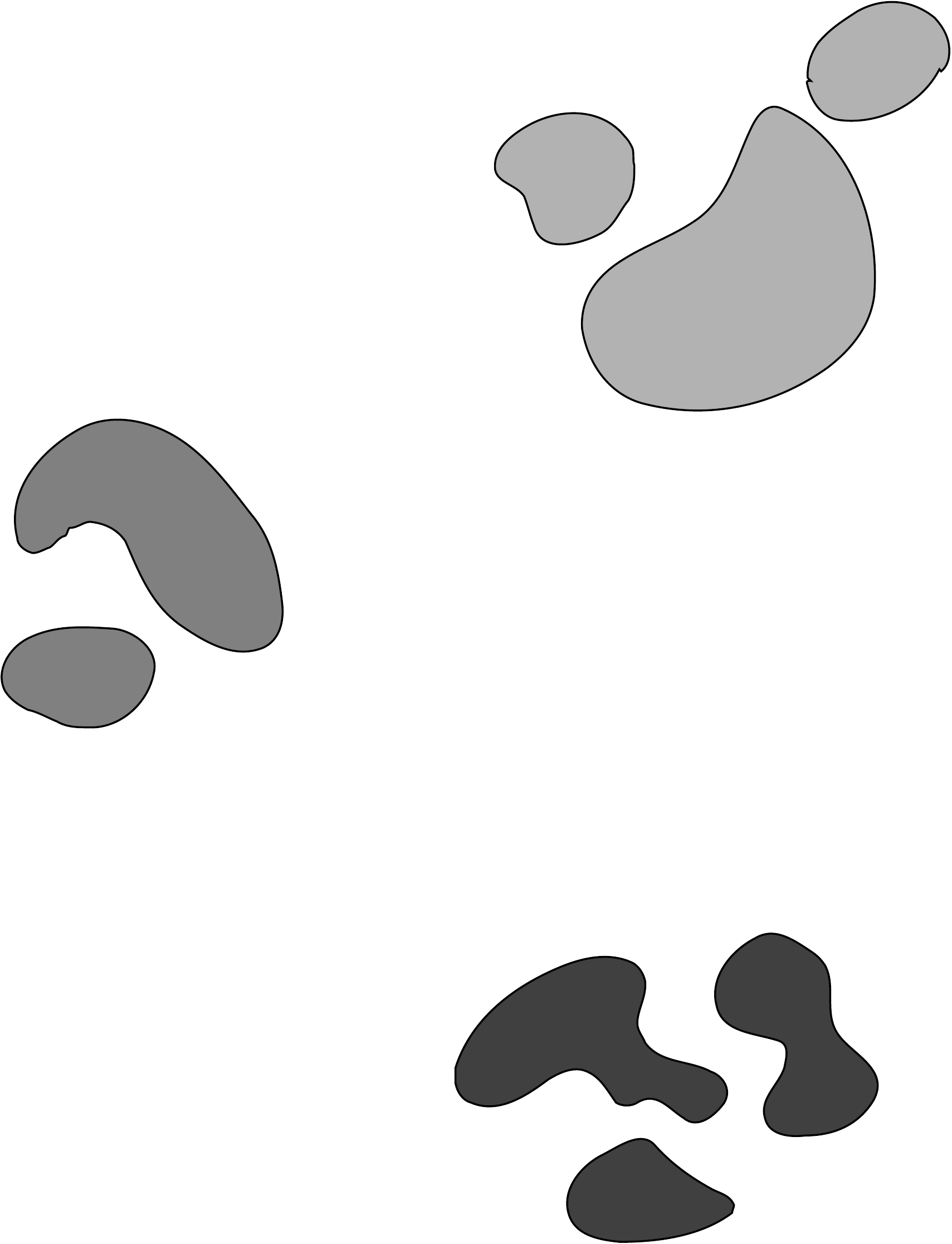} 
\hspace*{2cm}
\includegraphics[width=1.5cm]{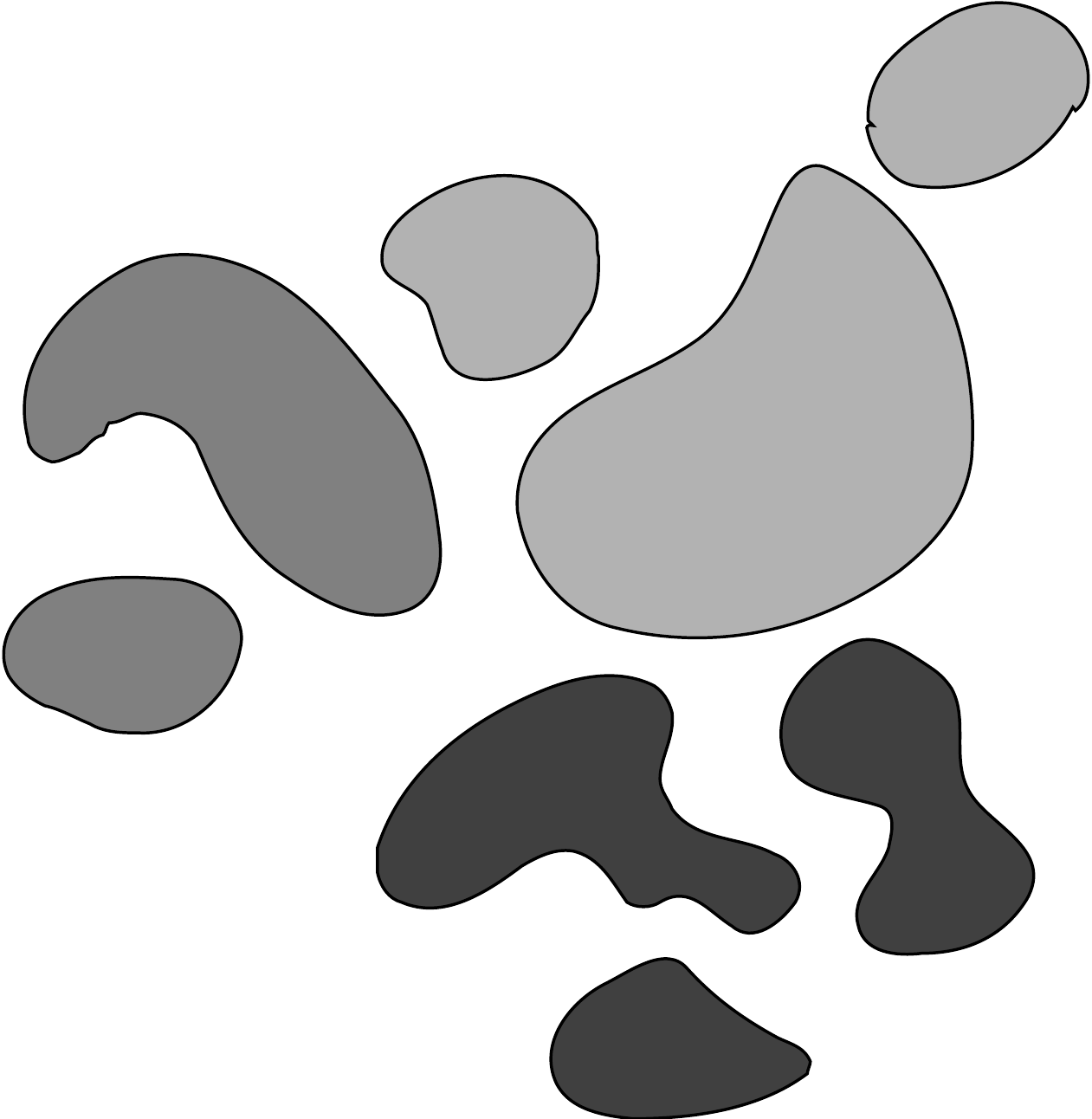}  
\end{center}  
a) \hspace{1.5in} b) \hspace {1.5in}

\caption{\label{fig.II4} Subconfigurations of objects.  In a), a configuration 
is formed from three groups of objects.  In b), the groups of objects have 
been moved closer; based on geometric position, the groups are no longer 
clearly distinguished.}  
\end{figure}

Third, we seek a measure to distinguish how significant are objects 
within the configuration and to identify those that are mainly outliers.  
This would provide for a configuration a {\it hierarchical structure for the 
objects}, indicating which objects are most central to the configuration 
and which are less {\it positionally significant}.  For example, in 
Figure~\ref{fig.II3}, the position of object $\gW_1$ makes it more 
important for the overall configuration in b) than in a), where it is more 
of an \lq\lq outlier.\rq\rq\,  A small movement of $\gW_1$ in 
a) would be less noticeable and have a smaller effect to the overall 
configuration than in b).  By having a smaller effect we mean that the 
deformed configuration could be mapped to the original by a 
diffeomorphism which has smaller local distortions near the 
configuration in the case of a) versus b). Finally, there is the question of 
whether there are numerical invariants which can be used to determine 
when there are identifiable subconfigurations.   An example of 
this is seen in Figure~\ref{fig.II4}.  
\par

\par

In Sections \ref{S:secII.Link.Flow} and \ref{SII:sec.int} we consider the 
\lq\lq linking flow\rq\rq associated to the linking structure. The 
nonsingularity 
of the linking flow is guaranteed in \S \ref{S:secII.Link.Flow} by \lq\lq 
linking 
curvature conditions\rq\rq\ \, on the linking functions, given in a form 
which extends that given in \cite[Thm 2.5]{D1} for the radial flow.  Next, 
in \S \ref{SII:sec.int}, we use this linking flow to identify both the 
internal regions of neighboring objects and external regions shared by 
them, which are the \lq\lq linking neighborhoods\rq\rq \,between objects.  
Then, in \S \ref{SII:SkelLnkInt}, we show how to compute integrals over
the boundaries of the objects or over general regions inside or outside the
objects as \lq\lq medial and skeletal linking integrals,\rq\rq which are 
integrals 
defined on the skeletal sets in the interiors of the objects.  Lastly, in \S 
\ref{SecII:PosGeom}, we introduce and compute several 
\lq\lq volumetric--based\rq\rq\ numerical invariants that include 
measures of the relative closeness of neighboring 
objects and relative significance of the individual objects.  
We furthermore show how they may be computed from the medial/skeletal 
linking structure  via skeletal linking integrals.
\par
Then, in \S \ref{SecII:TierGrph.PxoxMatr} we will combine the invariants 
which measure these geometric features in two different ways.  One is to 
construct a \lq\lq proximity matrix\rq\rq which captures the closeness 
of all objects in the configuration and to which the Perron-Frobenius 
theorem can be applied, yielding a unique set of \lq\lq proximity 
weights\rq\rq assigned to the objects, measuring their overall closeness 
in the configuration.  The 
second is to construct a \lq\lq tiered graph structure,\rq\rq\, which is a 
graph with vertices representing the objects, edges between vertices of 
neighboring 
objects, and values of significance assigned to the vertices, and closeness 
assigned to the edges.  Then, as thresholds for closeness and significance 
vary, the resulting subgraph satisfying the conditions will exhibit the 
central objects (and outliers) of the configuration, various subgroupings 
of objects and 
a hierarchical ordering of the relations between objects.    The 
skeletal linking structure also allows for the comparison and statistical 
analysis of collections of objects in $\R^2$ and $\R^3$, extending the 
analyses given in earlier work for single objects.  \par

\section{Linking Flow and Curvature Conditions}
\label{S:secII.Link.Flow}
\par
We model a configuration of objects in 2D or 3D images by a collection of 
regions $\boldsymbol \Omega = \{ \gW_i\}$ in either $\R^2$ or $\R^3$, with each $\gW_i$ 
modeling one of the objects, whose boundaries $\cB_i$ may share common 
boundary regions (see e.g. \cite[Fig. 1 and 2]{DG1}) and along their edges 
there are singularities of generic type corresponding to whether the 
objects are flexible or rigid (see e.g. \cite[Fig. 5 and 6]{DG1}). 
\subsection*{Medial/Skeletal Linking Structures}
\par
We recall from \cite{DG1} the definition of a skeletal linking  structure 
$\{(M_i, U_i, 
\ell_i)\}$ for the multi-object configuration $\{ \gW_i\}$.  First 
of all, it consists of skeletal structures $\{(M_i, U_i)\}$ for each 
$\gW_i$, where each $M_i$ is a stratified set in $\gW_i$ and $U_i$ is a 
multi-valued vector field on $M_i$ whose vectors end at boundary points 
in $\cB_i$ (see\cite{D3} for 2D or 3D regions, or more generally 
\cite{D2}).  By $M_i$ being a stratified set we mean for regions in $\R^2$ 
it is a disjoint collection of smooth curve segments ending at branching or 
end points, and for $\R^3$, it is a disjoint collection of smooth surface 
regions, curve segments and points, with the surfaces ending at the curves 
and points.  We may 
define a stratified set $\tilde M_i$ from $M_i$ by replacing points $x \in 
M_i$ by pairs $(x, U_i(x))$, where $U_i(x)$ varies over the multiple values 
of $U_i$ at $x$, with strata formed from strata $S_j$ of $M_i$ together 
with a choice of smoothly varying values $U_i(x)$ for $x \in S_j$.  The 
mapping $\pi : \tilde M_i \to M_i$ sending $(x, U_i(x)) \mapsto x$ sends 
the strata $\tilde M_i$ to strata of $M_i$.  This has the benefit of being 
able to consider multi-valued objects on $M_i$ as single-valued objects 
on $\tilde M_i$.  \par
We express $U_i = r_i\bu_i$, where $\bu_i$ are multi-valued unit vector 
fields on $M_i$.  In addition, the $\ell_i$ are multi-valued \lq\lq linking 
functions\rq\rq defined on $M_i$, and there are then defined the 
multi-valued \lq\lq linking vector fields\rq\rq  $L_i = \ell_i \bu_i$ on 
each $M_i$.  These become single valued on $\tilde M_i$.  \par 
There are additional conditions of \cite[Def. 3.2]{DG1} to be satisfied for 
$\{(M_i, U_i, \ell_i)\}$ to be the skeletal linking structure for $\bgW = \{ 
\gW_i\}$.  Conditions S1 - S3 concern a refinement of the stratification 
of $\tilde M_i$ and the differentiability properties of $\ell_i$ on the 
strata of the refinement.  Also, the conditions L1 - L4 concern the 
relations between the linking vector fields from different objects and the 
nonsingularity of the \lq\lq linking flows\rq\rq\, generated by the linking 
vector fields.  Regions defined by the linking flows are what we use to 
identify properties of the positional geometry of the configuration.  
\par
\subsection*{Nonsingularity of the Linking Flow} 
\par
The nonsingularity of the radial flow, which occurs within the regions, 
was established in \cite[\S 4]{D1} (and see \cite[\S 2]{D3}) using the 
radial and edge shape operators $S_{rad}$ and $S_{E}$, which are 
multi-valued operators defined on $M_i$, with $S_{rad}$ defined at all 
points except the edge points $\partial M_i$ of $M_i$ and $S_{E}$ defined 
at these points.  These capture the geometric properties of the radial 
vector field $U_i$ and play an important role in determining the local 
geometric properties of the boundary and the relative and global 
geometric properties of the region $\gW_i$ (see \cite{D2} and \cite[\S 3, 
4]{D3}).  Although their properties differ from those of the differential 
geometric shape operators appearing in differential geometry, their 
eigenvalues $\gk_{r\, i}$, the {\em principal radial curvatures} and 
generalized eigenvalues $\gk_{E\, i}$, the {\em principal edge curvatures}, 
play an equally important role.  \par
We next explain how they appear in the sufficient conditions we give for 
nonsingularity of the linking flow.  As 
well we give formulas for the evolution of the radial and edge shape 
operators under the linking flow.  \par
Recall \cite[(3.1)]{DG1}, the {\it linking flow} from $M_i$ is defined by 
\begin{equation}
\label{Eqn2.1.0}
\gl_{i}(x,t) = x + \chi_i(x,t)\bu_i(x)\, ,
\end{equation}
 where $\bu_i(x)$ ranges over all possible values and
\begin{equation}
\label{Eqn2.1}
\chi_i(x,t) = \left\{      
\begin{array}{lr}       
 2tr_i(x) &  \displaystyle 0 \leq t \leq \frac{1}{2}\\        2(1-t)r_i(x) + 
(2t-1)\ell_i(x) &  \displaystyle \frac{1}{2} \leq t \leq 1      
\end{array}    \right.  \, .
\end{equation}
For $0 \leq t \leq \frac{1}{2}$, this flow is the radial flow at twice the 
speed and it extends the radial flow to the exterior  of the regions for 
$\frac{1}{2} \leq t \leq 1$ and ends at the \lq\lq linking axis\rq\rq 
$M_0$.  We let $\gl_{i\, t}(x) = \gl_i(x ,t)$ for each $t$ and refer to 
$\gl_{i\, 1}$ as the {\em linking mapping} from strata of the refinement 
of $\tilde M_i$ to strata of the linking axis.  We refer to the 
combined union of the $\gl_{i\, t}$ for all $i$ by $\gl_{t}$.  
\par
To establish the conditions for the nonsingularity of the linking 
flow for the skeletal linking  structure $\{(M_i, U_i, \ell_i)\}$, we 
introduce the following two conditions:
\begin{enumerate}
\item ({\it Linking Curvature Condition} )  For all points $x_0 \in M_i 
\backslash \partial M_i$ and all values $U_i(x_0)$,
\begin{equation*}
\ell_i < \min \{ \frac{1}{\gk_{r\, j}}\} 
\end{equation*}
\noindent for all positive 
principal radial curvatures $\gk_{r\, j}$;
\item ({\it Linking Edge Condition} )  For all points  $x_0 \in 
\overline{\bdyM}$ (the closure of $\bdyM$),
\begin{equation*}
\ell_i < \min \{ \frac{1}{\gk_{E\, j}}\}
\end{equation*}
\noindent for all positive 
principal edge curvatures $\gk_{E\, j}\,$.
\end{enumerate}
In these conditions, the values of $\ell_i$ and either $\gk_{r\, j}$ or 
$\gk_{E\, j}$ are at the same point $x_0$ and for the same value of 
$U_i(x_0)$. \par
The nonsingularity of the linking flow (and the radial flow) for a 
skeletal linking structure is given by the following.


\begin{Thm}[Nonsingularity of the Linking Flow]
\label{Thm1} 
Let $\{(M_i, U_i, \ell_i)\}$ be a skeletal linking structure in $\R^2$ or 
$\R^3$ which satisfies: the Linking Curvature Conditions and Linking Edge 
Conditions on all of the strata of all $M_i$.  Then, the radial flow also 
satisfies the radial curvature and edge curvature conditions on each 
stratum.  Hence, the following properties hold. \flushpar
\begin{itemize}
\item[i)]  On each stratum $S_j$ of the refinement of $\tilde M_i$, the 
linking flow is nonsingular and remains transverse to the radial lines.  
\item[ii)]  Hence, the image $W_j$ of the linking map on $S_j$ is locally a 
smooth stratum of the same dimension and which may only have nonlocal 
intersections from distant points in $S_j$.  If there are no nonlocal 
intersections then $W_j$ is a smooth stratum.  
\item[iii)]  The image of a stratum $S_j$ of $\tilde M_i$ under the radial 
map is a smooth stratum of $\cB_i$ of the same dimension. 
\item[iv)]  For both flows, at points of the top dimensional strata, the 
backward projection along the lines of $L_i$ will locally map strata of 
$\cB_i$, resp. $M_0$, diffeomorphically onto the smooth part of $M_i$.  
\item[v)]  Thus, if there are no nonlocal intersections, each $\cB_i$ will 
be a piecewise smooth embedded surface.  
\end{itemize}
\end{Thm}
\par 
The proof of this theorem follows from Proposition 8.1 in \cite{DG} and 
its corollaries along with Theorem 2.5 of \cite{D3}; see also 
\cite[Chap. 6]{Ga}.  
\par

\subsection*{Evolution of the Shape Operators Under the Linking Flow} 
\par
We may translate the vectors $U_i$ along the lines of each $L_i$ to the 
level sets of the linking flow.  We may use these translated vectors as a 
radial vector field on the level set.  Hence, for each $1 \leq t \leq 1$ we 
may define corresponding radial shape operators $S_{rad\, t}$ on the level 
sets (curves or surfaces) $\cB_t = \gl_t(M_i)$ in a neighborhood of 
$\gl_t(x_0)$.  We can best relate them to the radial or edge shape 
operators on $M_i$\, in terms of their matrix reprsentations.  For $x_0 \in 
M_i \backslash \overline{\bdyM}$ and each choice of $U_i(x_0)$, there is a 
smooth stratum $M_{i\, j}$ of $M_i$ containing $x_0$ in its closure and 
which smoothly extends through $x_0$ and a smoothly varying value of 
$U_i$ defined in a neighborhood of $x_0$ extending $U_i(x_0)$.  We denote 
the tangent space to this stratum at $x_0$ by $T_{x_0}M_{i\, j}$.  For 
$S_{rad}$ we choose a basis $\bv$ for $T_xM_{i\, j}$, which is either a 
single vector $\bv = \{v_1\}$ for configurations in $\R^2$, or $\bv = 
\{v_1, v_2\}$ for $\R^3$.  We then let $\bv^{\prime}$ denote the image of 
$\bv$ under $d\gl_{i\, t}$ which is a basis for the tangent space to the 
level set $\cB_t$ at $d\gl_{i\, t}$.  At points $x_0 \in 
\overline{\bdyM_i}$ for $\R^3$, instead a nonzero vector $v_1 \in T_{x_0} 
\partial M_{i\, j}$ is completed to a basis using $U_i(x_0)$ for the source 
and the unit normal vector $\bn(x_0)$ to $M_{i\, j}$ for the target.  We 
denote the resulting matrix representation of $S_{E}$ by $S_{E\, \bv}$; 
but for $t > 0$, it evolves to also become radial shape operators $S_{rad\, 
t}$; and we use a basis $\bv^{\prime\prime}$, which is the image of the 
basis $\bv$ of $T_{x_0}\partial M_{i\, j}$ under $d\gl_t(x_0)$ with 
$\bn(x_0)$ adjoined.  
\par
\begin{Remark}
\label{Rem2.1}
\normalfont
In the 3D case $S_{\bv}$ and $S_{E\, \bv}$ are $2 \times 2$ matrices; 
while in the 2D case, $S_{\bv}$ is a $1 \times 1$ matrix formed from the 
single radial curvature $\gk_r$ (see e.g. Examples 2.3 and 2.4 in 
\cite{D3}).  
\end{Remark}
\par
Then, the evolved radial shape operators under the linking flow are given 
by the following.  

\begin{Proposition}[Evolution of the Shape Operators]  \hfill
\label{Prop.II3.5}
\par 
Suppose $x_0 \in M_i$ with value $U_i(x_0)$ (with a smooth value of 
$U_i$ in a neighborhood of $x_0$).  Provided the following conditions are 
satisfied, the linking flow is nonsingular and the evolved radial shape 
operator on the level surface $\cB_t = \gl_t(M_i)$ in a neighborhood of 
$\gl_t(x_0)$ is given by the following.
\begin{itemize}
\item[1)] For $x_0 \in M_i \backslash \overline{\bdyM}$, 
if $\frac{1}{\chi(t)}$ is not an 
eigenvalue of $S_{\bv}$ for $0 \leq t \leq 1$, then
$$  S_{\bv^{\prime}, t} \,\,  = \,\,   (I - \chi(t) S_{\bv})^{-1}S_{\bv} 
\, .  $$
\item[2)]  For $x_0 \in \overline{\bdyM}$, if $\frac{1}{\chi(t)}$ is not a 
generalized eigenvalue of $(S_{E, \bv}, I_{n-1,1})$ for $0 < t \leq 1$, then
$$  S_{\bv^{\prime\prime}, t} \,\,  = \,\,   (I_{n-1,1} - \chi(t) S_{E, 
\bv})^{-1}S_{E, \bv} \, .  $$
\end{itemize}
\end{Proposition}
\par
The derivation of these results can be found in \cite[\S 7]{DG}  and 
\cite[Chap. 6]{Ga} extending the results in \cite{D2}, or see \cite[\S 
2]{D3}.  
\vspace{2ex}

\subsubsection*{Shape Operators on the Boundary and  Linking Medial 
Axis} 
\par
As a consequence of the corollaries, we can deduce the shape operator for 
the linking axis $M_0$, and in a region where the {\em partial Blum 
condition} is satisfied, the differential geometric shape operator on the 
boundary.  First, for the boundary, it is reached at $t = \frac{1}{2}$.  If for 
$x_0 \in M_i$ the radial vector field is orthogonal to $\cB_i$ at the point 
$x^{\prime} = \gl(x_0, \frac{1}{2})$, then we say that the skeletal 
structure $(M_i, U_i)$ satisfies the {\em partial Blum condition} at 
$x^{\prime}$.  If it satisfies the partial Blum condition in a neighborhood 
of a smooth point of $\cB_i$, then $S_{\bv^{\prime}, \frac{1}{2}}$ in 
Proposition \ref{Prop.II3.5} gives the differential geometric shape 
operator for $\cB_i$ at $x^{\prime}$, and hence completely describes the 
local geometry of the boundary at $x^{\prime}$.  This result and its 
consequences follow from \cite[\S 3]{D2} and also see \cite[\S 3]{D3}. 
\par
If instead we consider the image $x^{\prime} \in M_0$ of $x_0 \in M_i$ 
under the linking flow, then to such a point there is the corresponding 
point $x^{\prime\prime} = \gl(x_0, \frac{1}{2}) \in \cB_i$.  We then have a 
value of a radial vector field $U_0 = -(\ell_i - r_i)\bu_i$ at $x^{\prime}$, 
with $\ell_i$, $r_i$, and $\bu_i$ associated to the value $L_i(x_0)$.  This 
vector at the point $x^{\prime}$ ends at $x^{\prime\prime}$.  
This defines a multi-valued vector field $U_0$ on $M_0$.  Thus, we can 
view $(M_0, U_0)$ as a skeletal structure for the exterior region.  We can 
determine the corresponding radial or edge shape operator at $x^{\prime}$ 
by the following (see \cite[Cor. 8.7]{DG}).  
\begin{Corollary}  
\label{Cor.II3.7}
If $x^{\prime} \in M_0$ is as in the above discussion, then the radial shape 
operator for the skeletal structure $(M_0, U_0)$ at $x^{\prime}$ is given 
by either: if $x_0$ is a non-edge closure point, then with the notation of 
Proposition \ref{Prop.II3.5}, 
$$  S_{\bv^{\prime\prime}, t} \,\,  = \,\,  - (I - \ell_i S_{\bv})^{-1}S_{\bv} 
\, ;  $$
or if $x_0$ is an edge closure point, then 
$$  S_{\bv^{\prime\prime}, t} \,\,  = \,\,  - (I_{n-1,1} - \ell_i 
 S_{E, \bv})^{-1}S_{E, \bv} \, .  $$
\end{Corollary}
\par
This follows because the associated unit vector field at $x^{\prime}$ is 
$\bu_0 = -\bu_i$.  Thus, the radial shape operator for $(M_0, U_0)$ at 
$x^{\prime}$ is the negative of that for the stratum of $M_0$, viewed as a 
level set of the linking flow from $x_0 \in M_0$.  Hence, by Proposition 
\ref{Prop.II3.5} we obtain the result. 

\section{Positional Properties of Regions Defined Using the Linking Flow} 
\label{SII:sec.int}
\par
We next consider how the medial/skeletal linking structure $\{(M_i, 
U_i, \ell_i)\}$ for a multi-object configuration $\bgW = \{ \gW_i\}$ in 
$\R^2$ or $\R^3$ allows us to introduce numerical measures capturing 
various aspects of the object\rq s positions within the configuration.  
There are two possibilities for this.  One is to base the numerical 
quantities on geometric properties of the boundaries of the objects.  The 
second is to use instead volumetric measures for subregions of the 
objects and identified regions in the external complement which capture 
positional information about the objects.  
\par
\begin{figure}[!t]
\begin{center} 
\includegraphics[width=6.0cm]{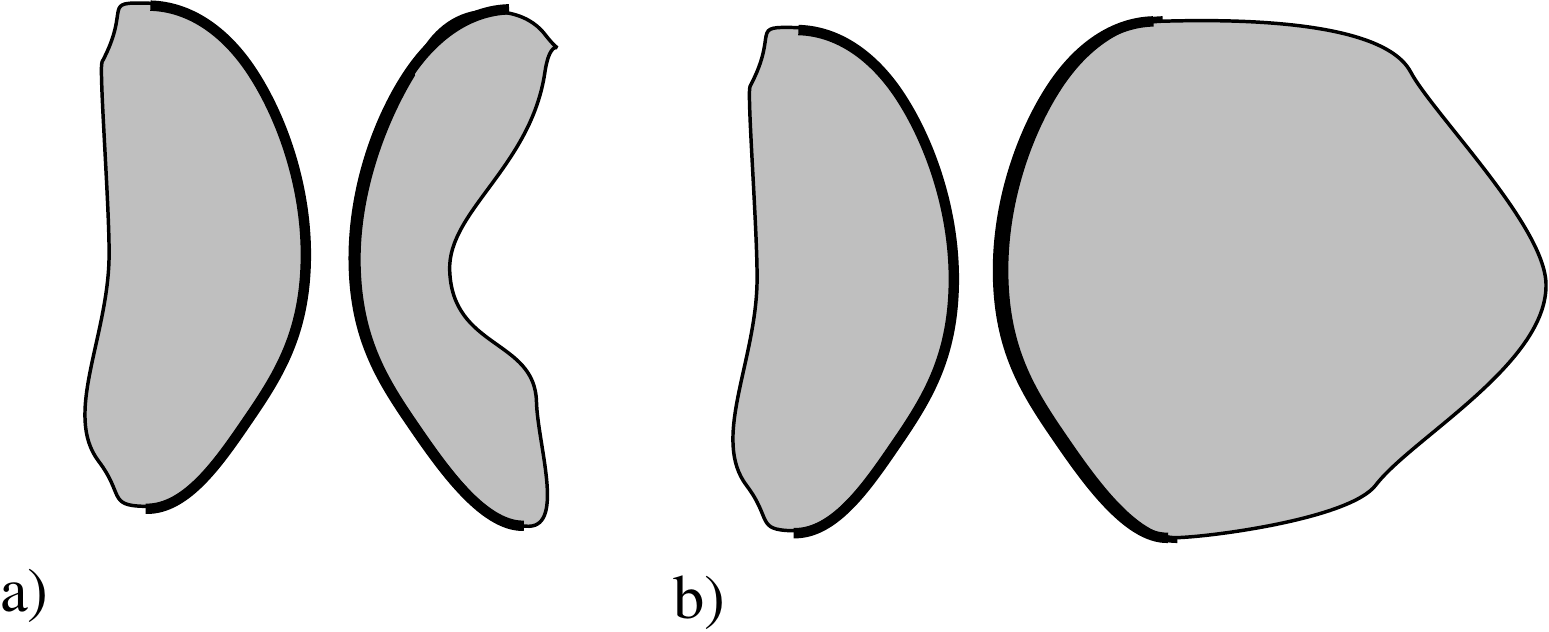}  
\end{center}  
\caption{\label{fig.III1} Pairs of regions in a) and b) both show the same 
boundary regions to each other as indicated by the dark curves.  However, 
in b) the region on the right has substantially more volume than does the 
region on the right  in a). This is not captured by neighboring 
boundary information; however, it is captured using volumetric measures 
defined from the linking structure.}  
\end{figure}
\par
The problem with the first choice is that two regions may show the same 
boundary region to each other even though there are other pairs of regions 
showing the same boundaries to each other but which may have completely 
different shapes and volumes, as shown in Figure \ref{fig.III1}.  The 
alternative is to keep track of the relation between all points of all 
boundaries of each pair of regions.  However, this involves an enormous 
redundancy in the data structure.  The skeletal linking structure avoids 
this redundancy, allowing additional geometric information to be 
computed directly from the linking structure.  We also shall see in 
\S \ref{SII:SkelLnkInt} that for general skeletal linking structures, using 
volumetric measures to capture positional information will allow us to 
express these numerical quantities as integrals over the internal skeletal 
sets $M_i$ of appropriate mathematical quantities derived from the 
linking structure .  \par
Because finite volumetric measures will require bounded regions in the 
complement, we will first consider regions defined in the unbounded case 
and then introduce bounded versions.  \par
\subsection*{Medial/Skeletal Linking Structures in the Unbounded Case}  
\par
\begin{figure}[!t]
\begin{center}
\includegraphics[width=4.2cm]{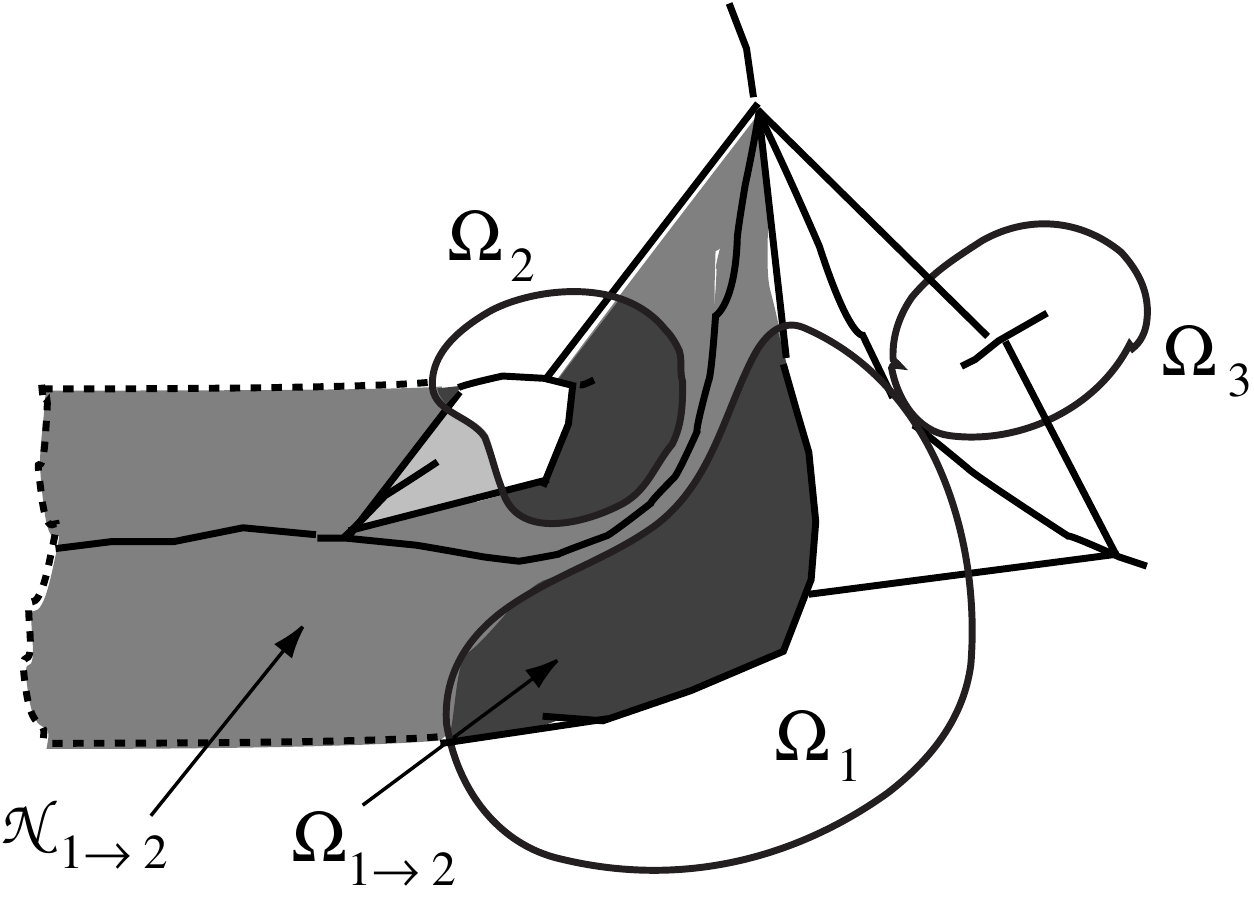} 
\hspace*{0.2cm}
\includegraphics[width=4.3cm]{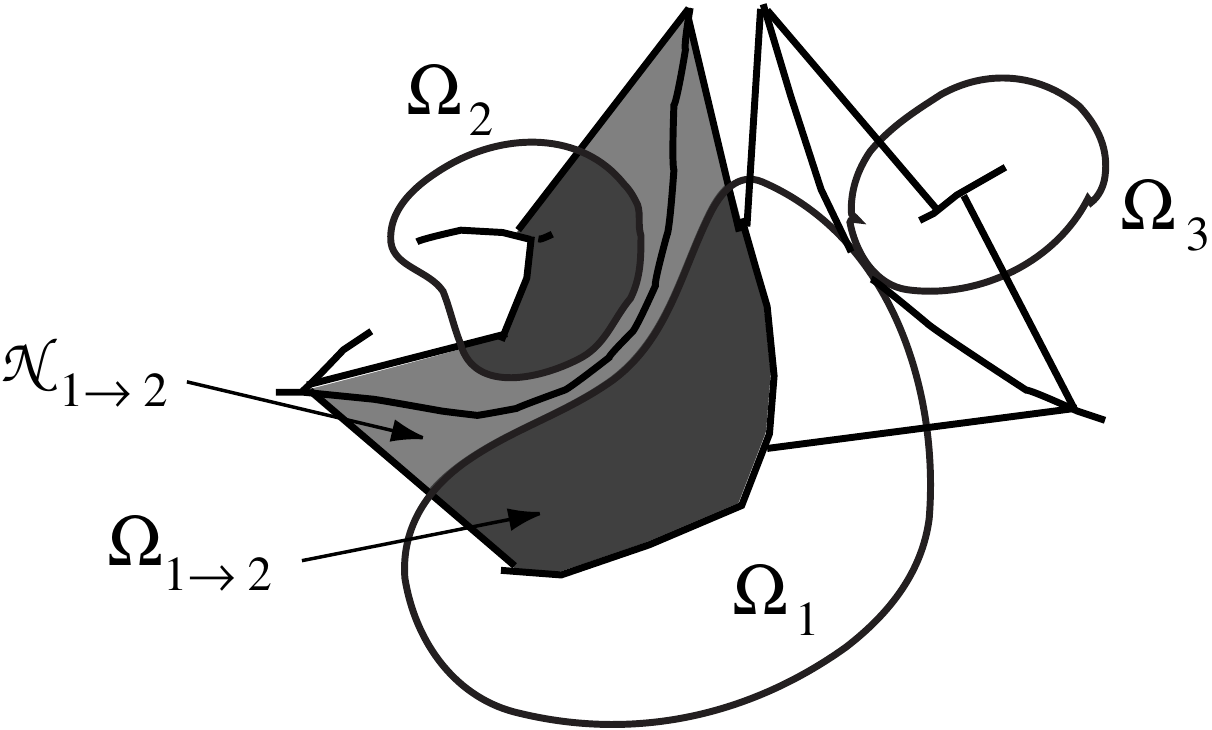}
\end{center}
a) \hspace{1.5in} b) \hspace {1.5in}
\caption{\label{fig.II4.1} Configuration of three regions with portions of 
the regions $\gW_1$ and  $\gW_2$ which are linked to each other (darkly 
shaded regions are parts of $\gW_{1 \to 2}$ and $\gW_{2 \to 1}$), and 
their linking neighborhoods (grey shaded regions are parts of $\cN_{1 \to 
2}$ and $\cN_{2 \to 1}$).  Then, $\cB_{1 \to 2}$ is the portion of the 
boundary $\cB_{1}$ where $\cN_{1 \to 2}$ meets $\gW_{1 \to 2}$, while 
$\cR_{1 \to 2}$ is the union of the two regions $\cN_{1 \to 2}$ and 
$\gW_{1 \to 2}$.  Note that in the unbounded case in a), much of the 
linking in the infinite region occurs between small parts of $\gW_1$ and 
$\gW_2$, and this would not occur for a bounded linking structure in a 
bounded region as in b) where a threshold is imposed.}
\end{figure}

 
We begin by considering regions $\gW_i$ and $\gW_j$ modeling objects in 
the configuration that are linked via the linking structure and 
identifying regions using the linking flow $\gl_{i}$ on each $\gW_i$.  We 
recall in \cite{DG1} that $\gW_i$ and $\gW_j$ are said to be 
\lq\lq linked\rq\rq\, if there are strata $S_{i\, k}$ in $M_i$ and $S_{j\, 
k^{\prime}}$ in $M_j$ which map to the 
same stratum in $M_0$ under the linking flow.  We then introduce the 
following regions as illustrated in Figure~\ref{fig.II4.1}. \par
\vspace{2ex}
\flushpar
{\it Regions Defined by the Linking Flow:} \par
\vspace{1ex}
\begin{itemize}
\item[i)] $M_{i \to j}$ will denote the union of the strata of $\tilde M_i$ 
which are linked to strata of $\tilde M_j$, and we refer to it as the {\it 
strata where $M_i$ is linked to $M_j$} (the strata being in $\tilde M_i$ 
indicate on which \lq\lq side\rq\rq of $M_i$ the linking occurs).  
\item[ii)] $\gW_{i\to j} = \gl_{i}(M_{i\to j} \times \left[0, 
\frac{1}{2}\right])$ denotes the {\it region of $\gW_i$ linked to $\gW_j$}.   
\item[iii)]  $\cN_{i\to j} = \gl_{i}(M_{i\to j} \times \left[\frac{1}{2}, 
1\right])$ denotes the {\it linking neighborhood} of $\gW_i$ linked to 
$\gW_j$.  
\item[iv)]  $\cB_{i \to j} = \gW_{i \to j} \cap \cN_{i \to j}$ is the {\it 
boundary region of $\cB_i$ linked to $\cB_j$}. 
\item[v)]  $\cR_{i \to j} = \gW_{i \to j} \cup \cN_{i \to j}$, is the {\it 
total region for $\gW_i$ linked to $\gW_j$}.
\end{itemize} 
\par 
\par
Then, we make a few simple observations.  First, $\cN_{i \to j} \cap 
\cN_{j \to i}$ will consist of the strata of the linking axis $M_0$
 where the linking between $\gW_i$ and $\gW_j$ occurs; and second, the
 regions for a fixed $i$ but different $j$ may intersect on the images
 under the linking flow of strata where there is linking between $\gW_i$
 and two or more other objects.  \par
Next, strata of $M_i$ may involve \lq\lq self-linking,\rq\rq\, which 
means different strata of $\tilde M_i$ may be linked to each other\, (see 
e.g. \cite[Fig. 9]{DG1}).  We will still use the notation $M_{i \to i}$, 
$\gW_{i \to i}$, 
etc. for the strata, regions etc. involving self-linking.  Then, 
$\cN_{i \to i}$ will intersect $\cN_{i \to j}$ on strata where \lq\lq 
partial linking\rq\rq\, occurs.  \par
Finally the remaining strata in $\tilde M_i$ lie in $M_{i\, ,\infty}$, which 
is introduced in \cite{DG1} and consists of the union of strata which are 
unlinked.  By property $L_4$ in \cite[Def. 3.2]{DG1}, the 
radial flow from the union of the $M_{i\, ,\infty} \times (0, \infty)$ 
defines a diffeomorphic parametrization of the complement of the regions 
reached by the linking flow.  We may flow at twice the radial flow speed 
to agree with the linking flow on the rest of $\tilde M_i$.  We still refer 
to this completed flow as the linking flow, and then denote the 
corresponding regions for $M_{i\, ,\infty}$ by: $\gW_{i\, ,\infty}$, 
$\cN_{i\, ,\infty}$,  $\cB_{i\, ,\infty}$ and $\cR_{i\, ,\infty}$. \par
Then, we have the decompositions 
\begin{equation}
\label{EqnII4.2}
\gW_i \,\, =\,\,  (\cup_{j \neq i} \gW_{i \to j}) \cup \gW_{i \to i} \cup 
\gW_{i\, ,\infty}
\end{equation}
\noindent with
\begin{equation*} (\cup_{j \neq i} \gW_{i \to j} 
\cup \gW_{i \to i}) \cap \gW_{i\, ,\infty} \,\, =\,\,  \emptyset \, ,
\end{equation*} 
but the various $\gW_{i \to j}$ and/or $\gW_{i \to i}$ may have 
non-empty intersections, as explained above.  There are analogous 
decompositions for $\tilde M_i$ and $\cB_i$.  Also, we denote the {\it 
total linking neighborhood} by $\cN_i = \cup_{j \neq i} \cN_{i \to j}$.
Then, $\cN_i \cup \cN_{i \to i} \cup \cN_{i\, ,\infty}$ is the {\it total 
neighborhood of $\gW_i$} (in the complement of the configuration), whose 
interior consists of points external to the configuration which are closest 
to $\gW_i$.  \par 
We let $\cB_{i\, ,0}$ denote the portion of 
the boundary $\cB_i$ that is not shared with any other region.  We recall 
that on 
the strata of $\tilde M_i$ corresponding to those in $\cB_i\backslash 
\cB_{i\, ,0}$, the linking flow is constant in $t$ for $\frac{1}{2} \leq t 
\leq 1$.  Hence, $\cN_i$ at these points only consists of boundary points 
of $\cB_i\backslash \cB_{i\, ,0}$.  Off of these points we can describe the 
structure of $\cN_i$ using the linking flow. 
We summarize the  consequences of the properties of the linking flow (see 
\cite[Cor. 9.2]{DG}).  
\par
\begin{Corollary}
\label{CorII4.1}
For a multi-object configuration $\bgW = \{ \gW_i\}$ with skeletal 
linking structure $\{(M_i, U_i, \ell_i)\}$, there are the following 
parametrizations for each region associated to $\gW_i$ by the portions of 
the level sets of the linking flow in those regions:
\begin{itemize}
\item[1)]  $\gW_i \backslash M_i$ is parametrized by the level sets of the 
linking flow for $0 < t \leq \frac{1}{2}$;
\item[2)] $\cN_i \backslash (\cB_i \backslash \cB_{i\, ,0})$ is 
parametrized by the level sets of the linking flow for $\frac{1}{2} \leq t 
\leq 1$;
\item[3)] $\cN_{i \to i}$ is parametrized by the level sets of the linking 
flow for $\frac{1}{2} \leq t < 1$; and 
\item[4)] $\cN_{i\, ,\infty}$ is  parametrized by the level sets of the 
radial flow for $1 \leq t < \infty$.
\end{itemize}
\end{Corollary}


\subsection*{Medial/Skeletal Linking Structures for the Bounded Case}
 \par
As we have mentioned, because the regions are often unbounded, there is 
no meaningful numerical measure of their sizes.  We can overcome this 
problem for practical considerations by introducing a bounded region 
$\tilde \gW$ containing the configuration and such that its 
boundary $\partial \tilde \gW$ is transverse both to the 
stratification of $M_0$ and to the linking vectors on $M$.  In \cite{DG1}, 
we describe a number of different possibilities for obtaining bounded 
regions including: a bounding box or bounding 
convex region, the convex hull of the configuration, a natural or intrinsic 
bounding region, or a region defined by a user-specified threshold for 
linking, see, e.g., Figure \ref{figII5.9}.  

\par
Then, we can either truncate the linking vector field, or define it on all of 
$M_i$ for all $i>0$ by defining it on  $M_{\infty}$ and subsequently 
refining the stratification so that the linking vector field ends at 
$\partial \tilde \gW$ on appropriate strata.   This maintains the 
nonsingularity of the linking flow as we are merely either reducing 
$\ell_i$ or defining $L_i$ on $M_{i\,, \infty}$. 
\par
Thus, we have the corresponding properties from Corollary \ref{CorII4.1}, 
except that for properties 2), 3), and 4) the linking flow and corresponding 
regions $\cN_{i \to j}$, $\cN_{i \, ,\infty}$, and $\cR_{i \to j}$ may only 
extend to $\partial \tilde \gW$.  These regions, which are now compact, 
are obtained from the unbounded regions by intersecting them with $\tilde 
\gW$.  We refer to this as the bounded case, referring the reader \cite[\S 
3]{DG1} for more details.  Moreover, we still obtain analogous formulas for 
the evolution of the radial shape operators for those level sets of the 
linking flow while they remain within $\tilde \gW$.   \par


\subsection*{Relevance of the Linking Regions for Positional Geometry}
 \par
The regions capture various aspects of the positional geometric 
properties 
of the configuration.  Two regions $\gW_i$ and $\gW_j$ are neighbors if
the regions $\gW_{i \to j}$, $\cB_{i \to j}$, $\cN_{i \to j}$, etc are 
nonempty, as are the corresponding $\gW_{j \to i}$, etc.  Then, the $\cB_{i 
\to j}$ and $\cB_{j \to i}$ represent the boundary regions \lq\lq 
between\rq\rq these neighbors.  Moreover, the $\gW_{i \to j}$ and 
$\gW_{j \to i}$ represent the internal portions of the regions which are 
\lq\lq closest\rq\rq to the neighbors.  These can be compared to the 
linking neighborhoods $\cN_{i \to j}$ and $\cN_{j \to i}$ to see how close 
the neighbors are.  The larger the linking neighborhoods are compared to 
the internal neighboring regions the further away are the neighboring 
regions.  If one region has large linking neighborhoods relative to all of its 
neighbors, then it plays a less significant positional role for the 
configuration.  This perspective will lead us in \S \ref{SecII:PosGeom} to 
introduce volumetric invariants of these regions which capture this 
positional geometry.  Before doing so, we next explain how numerical 
volumetric invariants can be obtained from the linking structures using 
\lq\lq linking integrals\rq\rq on the skeletal sets of the regions. 
\par
\par

\section{Global Geometry via Medial and Skeletal Linking Integrals} 
\label{SII:SkelLnkInt}
\par
We now will use the associated regions we have introduced for a 
configuration via a skeletal linking structure to define quantitative 
invariants measuring positional geometry for the configuration.  We will 
do so in terms of integrals which are defined on the internal skeletal sets.  
We begin by defining these integrals, and then we give a number of 
formulas for integrals of functions on the regions or boundaries of the 
configuration in terms of these integrals on the internal skeletal sets 
(see \cite[\S 10]{DG} and for 2D and 3D single regions \cite[\S 3.4]{D5}). 
\par 
In practice, skeletal structures have been modeled discretely, as can be 
skeletal linking structures.  For these the integrals then can be discretely 
approximated from the linking structure to compute the appropriate 
numerical invariants. \par

\subsection*{Defining Medial and  Skeletal Linking Integrals}
\par
We begin by considering a medial or skeletal linking structure $\{(M_i, 
U_i, \ell_i)\}$ for the configuration $\bgW = \{ \gW_i\}$ in 
$\R^2$ or $\R^3$.  We again let $M = \coprod_{i > 0} M_i$ denote the 
disjoint 
union of the $M_i$ for each region $\gW_i$ for $i > 0$.   
Each $M_i$ has its double $\tilde M_i$, so we introduce the double $\tilde 
M$ for the configuration by $\tilde M = \coprod_{i > 0} \tilde M_i$.  For 
each $i > 0$, there is a canonical finite-to-one projection $\pi_i : \tilde 
M_i \to M_i$, mapping $(x, U(x)) \mapsto x$.  The union of these defines a 
canonical projection $\pi : \tilde M \to M$, such that $\pi | \tilde M_i = 
\pi_i$ for each $i > 0$.  \par
We will define the skeletal integral on $\tilde M$ for a multi-valued 
function $g : M \to \R$, by which we mean for any $x \in M_i$, $g$ may 
have a different value for each different value of $U_i$ at $x$.  Such a $g$ 
pulls-back via $\pi$ to a well-defined map $\tilde g : \tilde M \to \R$ so 
that $g \circ \pi = \tilde g$.  \par
We recall that by Proposition 3.2 of \cite{D4}, for each $i > 0$, there is a 
positive Borel measure $dM_i$ on $\tilde M_i$, which we call the {\it 
medial measure}. If we let $M_{i\, ,\ga}^{(j)}$, $j = 1, 2$, denote the 
inverse images of $M_{i\, ,\ga}$ under the canonical projection map $\pi_i 
: \tilde M_i \to M_i$, then each $M_{i\, ,\ga}^{(j)}$ is a copy of $M_{i\, 
,\ga}$, representing \lq\lq one side of $M_{i\, ,\ga}$\rq\rq\, with the 
smoothly varying value of $U_{i}$ associated to the copy.  For each copy 
we let $dM_{j\, i} = \rho_{j\, i} dV_i$.  Here $dV_i$ denotes the 
Riemannian length or area on $M_{i\, \ga}$, as $n = 1$, resp. $2$.  Also, 
$\rho_{j\, i} = \bu_{j\, i}\cdot \bn_{j\,i}$ where $\bu_{j\, i}$ is a value 
of the unit vector field corresponding to the smooth value of 
$U_{j\, i}$ for $M_{i\, ,\ga}^{(j)}$ and $\bn_{j\, i}$ is the normal unit 
vector pointing on the same side as $U_{j\, i}$.  \par
Then, the integral of the multi-valued function $g$ over $M_{i\, ,\ga}$ is, 
by definition, the sum of the integrals of the corresponding values of $g$ 
over each copy $M_{i\, ,\ga}^{(j)}$ with respect to the medial measure 
$dM_{j\, i}$.  It is shown in \cite{D4} that for continuous $\tilde g$, this 
gives a well-defined integral and this extends to integrals of \lq\lq Borel 
measurable functions\rq\rq  $\tilde h$ on $\tilde M_i$, which include 
piecewise continuous functions.  
Then, these distinct medial measures $dM_i$ on $\tilde M_i$ together 
define a {\it medial measure} $dM$ on $\tilde M$; and the integral of a 
Borel measurable multi-valued function $g$ on $M$
is defined to be
\begin{equation}
\label{EqnII5.1}
 \int_{\tilde M} g \, dM \,\, = \,\, \sum_{i> 0} \int_{\tilde M_i} \tilde{g}\, 
dM_i , 
\end{equation}
where each integral on the RHS is the integral of $\tilde g$ over $\tilde 
M_i$ with respect to the measure $dM_i$, and it can be viewed as an 
integral of $g$ over \lq\lq both sides of $M_i$.\rq\rq.  If the integral is 
finite then we say the function is \lq\lq integrable\rq\rq. \par
We refer to the integrals in (\ref{EqnII5.1}) as {\it medial 
or skeletal linking integrals}, depending on whether the linking structure 
is a Blum medial linking structure or a skeletal linking structure.  \par

\subsection*{Computing Boundary Integrals via Medial Linking Integrals} 
\par
We now show how, for a multi-object configuration with \lq\lq 
full\rq\rq\, Blum linking structure, we may express integrals of functions 
on the combined boundary $\cB$ as medial integrals.  We emphasize that 
the full Blum linking structure allows the Blum medial axis to extend up 
to the edge-corner points of the boundaries.  This does not alter the 
existence nor definition of the integrals, see \cite[\S 10]{DG}. \par 
First, we consider a Borel measurable and integrable function 
$g : \cB \to \R$
which is multi-valued in the sense that for any $k$-edge-corner point $x 
\in \cB$, 
$g$ may take distinct values for each region $\gW_i$, $i > 0$, containing 
$x$ on its boundary.  Thus, if $\cB_{i\, j}$ denotes the shared boundary 
region of $\cB_i$ and $\cB_j$, $g$ 
may take different values on $\gW_i$ and $\gW_j$.  For example, $\gW_i$ 
and $\gW_j$ may have different boundary properties such as densities 
measured by $g$.  \par
 By the integral of such a multi-valued function $g$ over $\cB$ we mean 
$$  \int_{\cB} g\, dV \,\, = \,\,  \sum_{i \neq j \geq 0} \int_{\cB_{i\, j}} 
g_{i\, j}\, dV $$
where $g_{i\, j}$ denotes the values of $g$ on $\cB_{i\, j}$ for $\gW_j$ 
and $dV$ denotes the Riemannian length (for 2D) or area (for 3D) for each 
$\cB_i$.
\par
Then, for the radial flow map $\psi_{i\, 1} : \tilde M_i \to \cB_i$, we 
define $\tilde g : M_i \to \R$ by $\tilde g = g \circ \psi_{i\, 1}$, where 
the value on $\cB_{i\, j}$ is the value associated to $\gW_i$.  Then, 
$\tilde g$ is a multi-valued Borel measurable function on $M_i$.  
We may compute the integral of $g$ over $\cB$ by the following result 
\cite[Thm 10.1]{DG}.
\begin{Thm}
\label{ThmII5.1}
Let $\bgW$ be a multi-object configuration with (full) Blum linking 
structure.  If $g : \cB \to \R$ is a multi-valued Borel measurable and 
integrable function, then
\begin{equation}
\label{EqnII5.2}
\int_{\cB} g\, dV \,\, = \,\, \int_{\tilde M} \tilde g \det(I - r_i S_{rad})\, 
dM\, 
\end{equation}
where $r_i$ is the radius function of each $\tilde M_i$. 
\end{Thm}
\par 
In the case of a skeletal structure, there is a form of Theorem 
\ref{ThmII5.1} which still applies.  For each region $\gW_i$, with $i > 0$, 
let $\tilde R_i$ denote a Borel measurable region of $\tilde M_i$ which 
under the radial flow maps to a Borel measurable region $ R_i$ of $\cB_i$.  
Let $R = \displaystyle \bigcup_{i} R_i$ and $\tilde R = \displaystyle 
\bigcup_{i} \tilde R_i$.  We suppose 
that the skeletal structure satisfies the \lq\lq partial Blum 
condition\rq\rq\, on $\tilde R$, by which we mean: for each $i$, the 
compatibility $1$-form $\eta_{U_i}$ vanishes on $\tilde R_i$ (recall this 
means that the radial vector $U_i$ at points of $x \in \tilde R_i$ is 
orthogonal to $\cB_i$ at the point where it meets the boundary).  Note 
that for a skeletal structure this forces $R$ to be contained in the 
complement of $\cB_{sing}$.  \par 
Then, there is the following analogue of Theorem \ref{ThmII5.1}.
\begin{Corollary}
\label{CorII5.1}
Let $\bgW$ be a multi-object configuration with skeletal linking 
structure which satisfies the partial Blum condition on the region $\tilde 
R \subset \tilde M$, with image $R$ under $\psi_1$.  If $g : R \to \R$ is a 
multi-valued Borel measurable and integrable function, then
\begin{equation}
\label{EqnII5.3b}
\int_{ R} g\, dV \,\, = \,\, \int_{\tilde R} \tilde g \det(I - r_i S_{rad})\, 
dM\, .
\end{equation} 
\end{Corollary}

\begin{Remark} \label{RemII5.5}
\normalfont
One may compute the length (2D) or area (3D) of $\cB$
by choosing $g \equiv 1$ in Theorem \ref{ThmII5.1}. 
There is an analogous result for a region $R \subset \cB$ which is the 
image of a region $\tilde R \subset \tilde M$ under the radial flow.  If the 
configuration is modeled by a skeletal linking structure which satisfies 
the partial Blum condition on $\tilde R$, then the length, resp. area, of 
$R$ is given by $\int_{\tilde R} \det(I - r_i S_{rad})\, dM$. 
\end{Remark}

\par 
\subsection*{Computing Integrals over Regions as Skeletal Linking 
Integrals} 
\par
Next, we turn to the problem of computing integrals over regions which 
may be partially or completely in the external region of the configuration.  
Quite generally we consider a Borel measurable and integrable 
scalar-valued function $g$ defined on $\R^2$ or $\R^3$, but only nonzero 
on a compact region.  We shall see that we can compute the integral of $g$ 
as an integral of an appropriate related function on the internal skeletal 
sets.  \par 
Since we are in the unbounded case, we first modify the skeletal linking 
structure by defining $\ell_i$ on $M_{i\, ,\infty}$ to be $\ell_i = \infty$.  
The linking flow on $M_{i\, ,\infty}$ is a 
diffeomorphism for $0 \leq t < \infty \,(= \ell_i)$, see 
\cite[Prop. 14.11]{DG}.  \par 
Next, we replace the linking flow by a simpler {\it elementary 
linking flow} defined by $\gl^{\prime}_t(x) = x + t \bu_i$, for $0 \leq t 
\leq \ell_i$ (or $< \infty$ if $\ell_i= \infty$).  The elementary linking 
flow is again along the radial lines determined by $L_i$; however, the rate 
differs from that for the usual linking flow.  This means that the level 
surfaces will differ, although the images of strata under the elementary 
linking flow agree with that for the linking flow.  In addition, as the 
linking flow is nonsingular, the linking curvature and edge conditions are 
satisfied.  Then, viewing the linking vector field as a radial vector field, 
the radial curvature and edge curvature conditions are satisfied, and hence 
imply the nonsingularity of the elementary linking flow.  
\par 
Then, using the elementary linking flow, we can compute the integral of 
$g$ as a skeletal linking integral.  We define a multi-valued function 
$\tilde g$ on $M$ as follows: for $x \in M_i$ with associated smooth value 
$U_i$ and linking vector $L_i$ in the same direction as $U_i$ (so $(x, U_i) 
\in \tilde M_i$),  
\begin{equation}
\label{EqnII5.6}
 \tilde g(x) \,\, \overset{def}{=} \,\, \int_{0}^{\ell_i} g (\gl^{\prime}_t(x)) 
\det(I - t S_{rad})\, dt  
\end{equation}
provided the integral is defined.  
Then, we have the following formula for the integral of $g$ as a skeletal 
linking integral \cite[Thm 10.6]{DG}. 
\begin{Thm}
\label{ThmII5.5}
Let $\bgW$ be a multi-object configuration in $\R^{n+1}$ $(n=1$ or $2)$ 
with a skeletal linking structure.  If $g : \R^{n+1} \to \R$ is a Borel 
measurable and 
integrable function which is zero off a compact region, then $\tilde g(x)$ 
is defined for almost all $x \in \tilde M$, it is integrable on $\tilde M$, 
and 
\begin{equation}
\label{EqnII5.7}
\int_{\R^{n+1}} g\, dV \,\, = \,\, \int_{\tilde M} \tilde g \, dM\, .
\end{equation} 
\end{Thm}
\par
\begin{Remark}
\label{RemII5.8}
\normalfont
If we compare this formula with that given for a single region in Theorem 
6.1 of \cite{D4}, we notice they have a slightly different 
form in that a factor of $\ell_i$ appears to be missing .  However, as 
noted in Remark 6.2 of that paper, it is possible to use a change of 
coordinates $t = \ell_i t^{\prime}$ to rewrite 
\begin{equation}
\label{EqnII5.8a}
\tilde g(x) \,\, = \,\, \ell_i \int_{0}^{1} g (x + t^{\prime} L_i) \det(I - 
t^{\prime} \ell_i S_{rad})\, dt^{\prime}\, , 
\end{equation}
so that the form of (\ref{EqnII5.8a}) agrees with the form given in 
\cite{D4}.  The apparent difference in form will also appear in all of the 
following formulas compared with the corresponding ones in \cite{D4}.
\end{Remark}
\par
\subsubsection*{Reducing to Integrals for Bounded Skeletal Linking 
Structures} \par
We may replace the unbounded skeletal structure by a bounded one and 
replace the integrals by integrals over bounded regions.  If $g$ is zero off  
the compact region $Q$, then we may find a compact convex region $\tilde 
\gW$ with smooth boundary containing both the configuration $\bgW$ and 
$Q$.  Then, we can modify the linking structure by reducing the $L_i$ so it 
is truncated where it meets $\partial \tilde \gW$, the boundary of $\tilde 
\gW$, and defining $L_i$ on $M_{i\, ,\infty}$ as the extensions of the 
radial vectors to where they meet $\partial \tilde \gW$.  Because the 
extended radial lines are transverse to 
$\partial \tilde \gW$, the new smaller values of $\ell_i$
 remain differentiable on the strata of each $M_i$.  Still letting the 
$\ell_i$ denote the new smaller values, the corresponding truncated 
vector fields will still be denoted by $L_i$.  Then the formula for the 
integral of $g$ is still given by (\ref{EqnII5.7}).  We shall assume we have 
chosen a bounded linking structure for the remainder of this section. 
\par
To simplify the statements for the remainder of this section, we shall use 
the notation for a region $Q \subset \R^2$ or $\R^3$: $\vol_2(Q) = 
\text{area}(Q)$ for $Q \subset \R^2$ or $\vol_3(Q) = \vol(Q)$ for $Q 
\subset \R^3$.  \par 
For each $x  \in \tilde M_i$, we let 
\begin{equation}
\label{EqnII5.13a}
 m_Q(x) \,\, =\,\, \int_{0}^{\ell_i} \chi_Q(x + t L_i(x)) \det(I - 
t S_{rad})\, dt  \, . 
\end{equation} 
We can view $m_Q(x)$ as a weighted $1$-dimensional measure of the 
length of the intersection of $Q$ with the linking line from $x$ 
determined by $L_i(x)$. Then, applying Theorem \ref{ThmII5.5} in the 
special case where $g \equiv 1$, we obtain an analogue of Crofton\rq s 
formula giving the area, resp. volume, of $Q$ as a skeletal integral of 
$m_Q$ using \cite[Cor. 10.9]{DG}. 
\begin{Corollary}[Crofton Type Formula]
\label{CorII5.5}
Let $\bgW$ be a multi-object configuration with skeletal linking 
structure in $\R^{n+1}$ for $n = 1$, or $2$.  Suppose $Q \subset \R^{n+1}$  
is a compact subset. Then
\begin{equation}
\label{EqnII5.13b}
\vol_{n+1}(Q) \,\, = \,\,  \int_{\tilde M} m_Q(x) \, dM\, .
\end{equation} 
\end{Corollary}
\par 
\subsubsection*{Decomposition of a Global Integral using the Linking 
Flow} 
\par  
We next decompose the integral on the RHS of (\ref{EqnII5.7}) into internal 
and external parts using the alternative integral representation of $\tilde 
g$ using the linking flow.  
We do so by applying the change of variables formula to relate the 
elementary linking flow $\gl^{\prime}$ with the linking flow $\gl$, both 
of which flow along the linking lines but at different linear rates.  \par
We define 
\begin{align}
\label{EqnII5.14}
  \tilde g_{int}(x) \,\, &= \,\, \int_{0}^{r_i } g(x + t\bu_i) 
\det(I - t S_{rad})\, dt\, \quad \text{ and } \quad \notag \\
\tilde g_{ext}(x) \,\, &= \,\, \int_{r_i }^{\ell_i} g(x + t\bu_i) \det(I - t 
S_{rad})\, dt. 
\end{align}
\par
\begin{figure}[!t]\centering
\includegraphics[width=7cm]{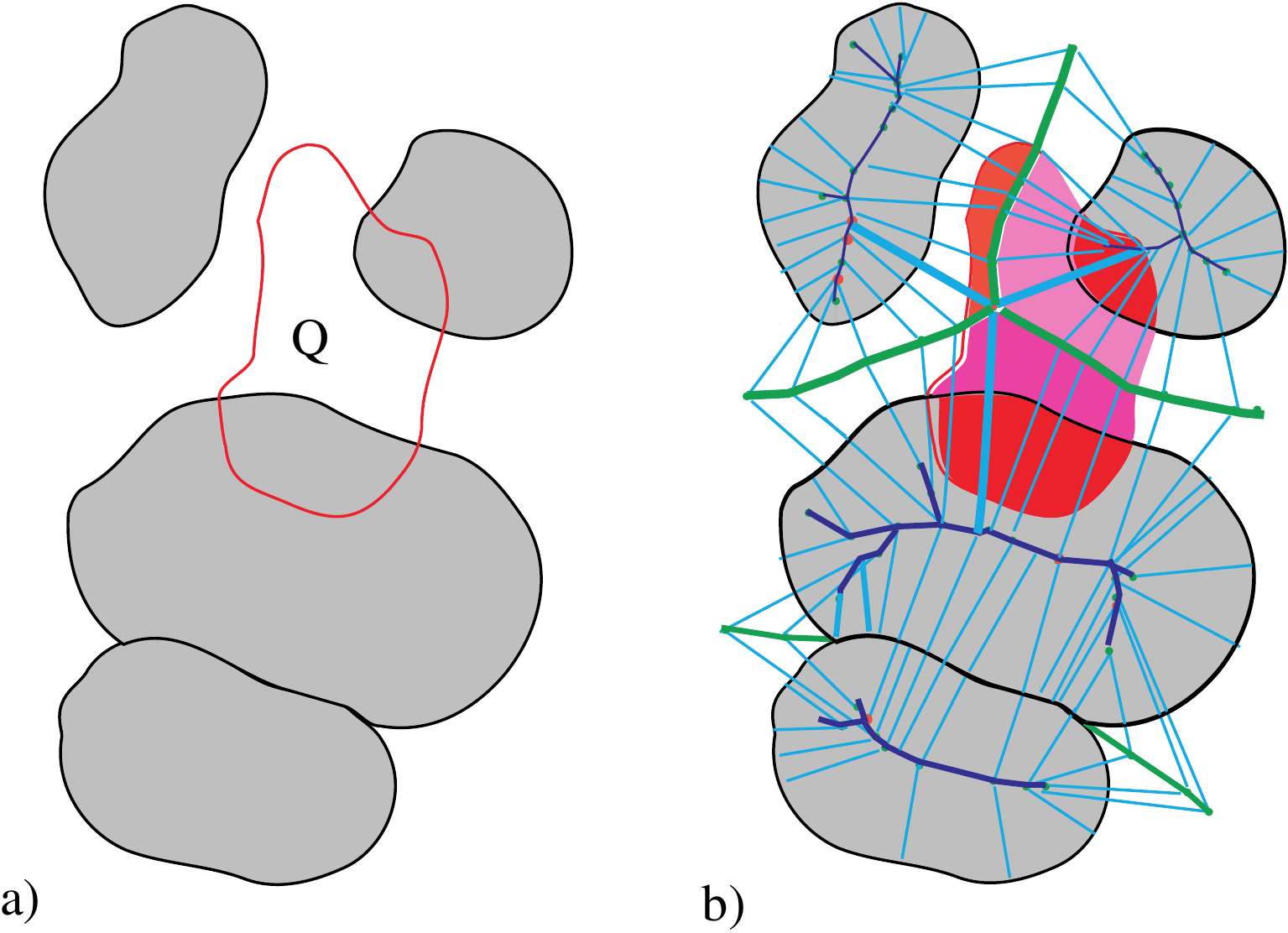}
\caption{ \label{figII5.8} The decomposition of the integral over a region 
$Q$ outlined in a) is given as the sum of integrals over regions in b)  
obtained by the subdivision of $Q$ by the linking axis and the three linking 
lines to the branch point of the linking axis.  Each $Q_{i\, j} \subset 
\cR_{i\to j}$ (or in general including $Q_{i\, \infty} \subset \cR_{i\, 
\infty}$) in the figure consists of the darker region inside the subregion 
$\gW_{i \to j}$ together with the portion of $Q$ in the linking 
neighborhood $\cN_{i\to j}$.  The integral can then be expressed by 
Corollary \ref{CorII5.6} as sums of internal and external integrals over 
the $M_{i\to j}$ and $M_{i\, \infty}$.}
\end{figure}

Then, we may decompose $\int g$ as follows, \cite[Cor. 10.10]{DG}.
\begin{Corollary}
\label{CorII5.6}
Let $\bgW$ be a multi-object configuration in $\R^2$ or $\R^3$ with a 
skeletal 
linking structure.  If $g : \R^{n+1} \to \R$ is a Borel 
measurable and integrable function for $n = 1$, resp. $2$, which equals 
$0$ off a compact set, then $\tilde 
g_{int}(x)$ and $\tilde g_{ext}(x)$ are defined for almost all $x \in \tilde 
M$, they are integrable on $\tilde M$, and 
\begin{equation}
\label{EqnII5.7b}
\int_{\R^{n+1}} g\, dV \,\, = \,\, \int_{\tilde M} \tilde g_{int} \, dM\, \, + 
\,\, \int_{\tilde M} \tilde g_{ext} \, dM \, ,
\end{equation} 
where 
\begin{equation}
\label{EqnII5.7c}
\int_{\tilde M} \tilde g_{int} \, dM \,\, = \,\, \sum_{i, j >0} \int_{ M_{i \to 
j}} \tilde g_{int} \, dM\, \, + \,\, \sum_{i >0} \int_{M_{i \, ,\infty}} \tilde 
g_{int} \, dM \, ,
\end{equation} 
with an analogous formula with $g_{int}$ replaced by $g_{ext}$ 
everywhere in (\ref{EqnII5.7c}).  
\end{Corollary}
The first integral on the RHS of (\ref{EqnII5.7b}) is the \lq\lq interior 
integral\rq\rq\, of $g$ within the configuration using the radial flow, and 
the second integral is the\lq\lq external integral\rq\rq\, computed using 
the linking flow outside of the configuration.  Then we may further 
decompose each of these integrals using (\ref{EqnII5.7c}) into integrals 
over the distinct linking regions as illustrated in Figure~\ref{figII5.8}.
\par
\par

\subsection*{Skeletal Linking Integral Formulas for Global Invariants} 
\par
We now express the areas (2D) or volumes (3D) of regions associated to 
the linking structure, which we introduced in \S \ref{SII:sec.int}, as 
skeletal linking integrals.  
We may apply the same reasoning as in Corollary \ref{CorII5.5}, using 
Corollary \ref{CorII5.6} to compute the volume of a compact 
2D or 3D region $Q$ as a sum of internal and external integrals.  
\par
For these calculations we will use the polynomial expression in $s$
\begin{equation}
\label{EqnII5.20a}
\cI(s) \,\, \overset{def}{=} \,\, \int_{0}^{s} \det(I - t S_{rad}) \, dt \, . 
\end{equation}
In the 2D case, $\cI(s) = s -\frac{\gk_r}{2} s^2$; and in the 3D case, 
$\cI(s) = s - H_{rad} s^2 + \frac{1}{3} K_{rad} s^3$, where $H_{rad} = 
\frac{1}{2}(\gk_{r\, 1} + \gk_{r\, 2})$ and $K_{rad} =\gk_{r\, 1} \cdot 
\gk_{r\, 2}$, for $\gk_{r\, i}$ the principal radial curvatures.    
\subsubsection*{Areas and Volumes of Linking Regions as Skeletal 
Integrals} 
\par
Then, we can express the areas or volumes of various linking regions as 
integrals of $\cI(s)$ for various choices of $s$. 
\begin{Corollary}
\label{CorII5.8}
Let $\bgW \subset \tilde \gW$ be a multi-object configuration with a 
bounded skeletal linking structure in $R^{n+1}$ for $n = 1$, resp. $2$.  
Then, the areas, resp. volumes, of linking regions in $R^{n+1}$ are given by 
the following:
\begin{align}
\label{EqnII5.20}
\vol_{n+1}(\cN_{i \to j}) \, &= \,  \int_{\tilde M_{i \to j}} \cI(\ell_i) - 
\cI(r_i)\, dM\, \,\, \notag ;\\
 \quad \vol_{n+1}(\gW_{i \to j}) \, &= \,  
\int_{M_{i \to j}} \cI(r_i)\, dM\, \notag ; \\
\vol_{n+1}(\cN_{i\, ,\infty}) \, &= \,  \int_{M_{i\, ,\infty}} \cI(\ell_i) - 
\cI(r_i)\, dM\, \,\, \notag ; \\
\quad \vol_{n+1}(\gW_{i\, ,\infty}) \, &= \,  
\int_{M_{i\, ,\infty}} \cI(r_i)\, dM\, .
\end{align} 
\end{Corollary}
As well as these formulas, we can compute the volumetric invariants of 
other linking regions such as $\cR_{i \to j}$, $\cN_{i}$, etc. using 
skeletal linking integrals of the polynomials $\cI(\ell_i)$ or $\cI(r_i)$.  
For example,
\begin{equation}
\label{EqnII5.20b}
 \vol_{n+1}(\cR_{i \to j}) = \int_{M_{i \to j}} \cI(\ell_i) \, dM \, , 
\end{equation}
 with an analogous formula for $\vol_{n+1}(\cR_{i \, ,\infty})$.  \par
As a consequence, we obtain generalizations of the classical formula of 
Weyl for \lq\lq volumes of tubes\rq\rq and Steiner\rq s formula for 
volumes of \lq\lq annular regions\rq\rq.  
\par
\begin{figure}[!t]
\includegraphics[width=8.7cm]{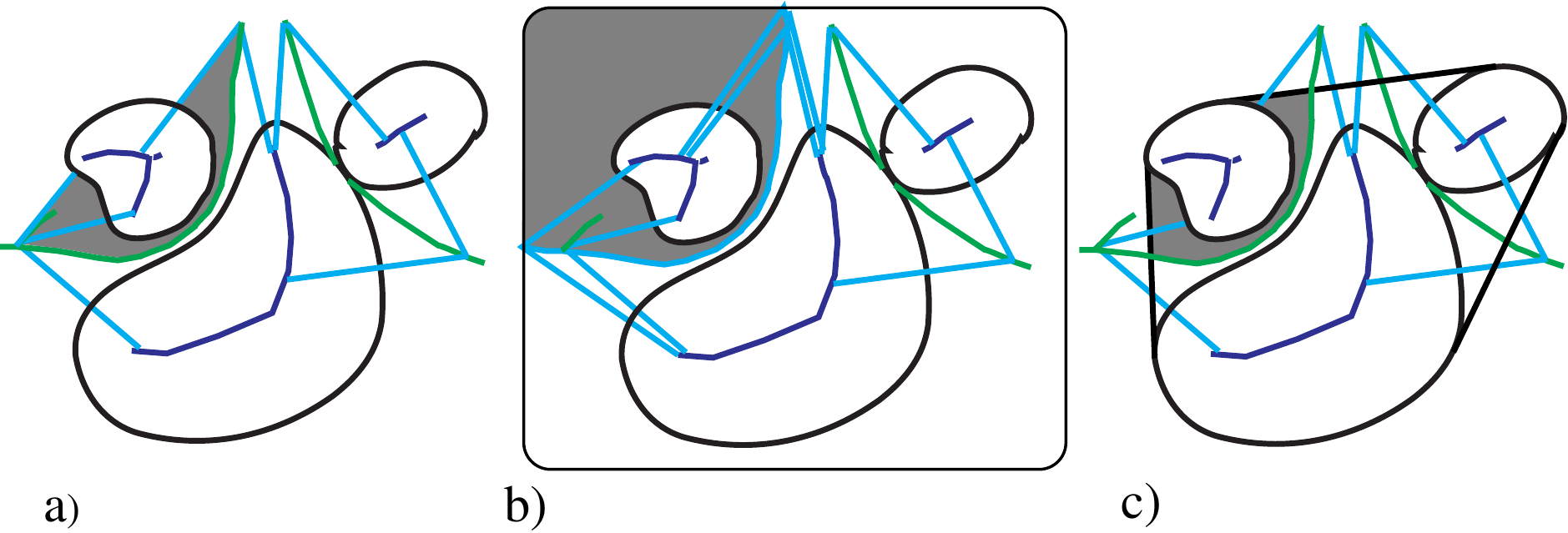}
\caption{\label{figII5.9}  Examples of a total neighborhood $\cN_i \cup 
\cN_{i\to i} \cup \cN_{i\, ,\infty}$ for an object $\gW_i$, to which the 
generalized Steiner\rq s formula applies: a) absolute threshold; b) 
bounding box; and c) convex hull.}
\end{figure} 
\par

\begin{Corollary}[Generalized Weyl\rq s Formula]
\label{CorII5.9}
Let $\bgW \subset \tilde \gW \subset \R^{n+1}$, for $n = 1$ or $2$, be a 
multi-object configuration with a bounded skeletal linking structure.  
Then, 
\begin{equation}
\label{EqnII5.21}
\vol_{n+1}(\gW_i) \,\, = \,\,  \int_{\tilde M_i} \cI(r_i)\, dM\, . 
\end{equation} 
\end{Corollary}
The sense in which this generalizes Weyl\rq s formula is explained for the 
case of a single region with smooth boundary in \cite[\S 6, 7]{D4}.  For 
Steiner\rq s formula, we note that as explained in \S \ref{SII:sec.int},  
$\cN_i \cup \cN_{i\to i} \cup \cN_{i\, ,\infty}$ represents the total 
neighborhood of $\gW_i$, which is the region about $\gW_i$ extending 
along the linking lines.  This is a generalization of an \lq\lq annular 
neighborhood\rq\rq about a region which depends on the specific type of 
bounding region (see Figure~\ref{figII5.9}). 
\begin{Corollary}[Generalized Steiner\rq s Formula]
\label{CorII5.10}
Let $\bgW \subset \tilde \gW \subset \R^{n+1}$, for $n = 1$ or $2$ be a 
multi-object configuration with a bounded skeletal linking structure.  
Then,
\begin{equation}
\label{EqnII5.22}
\vol_{n+1}(\cN_i \cup \cN_{i\to i} \cup \cN_{i\, ,\infty} ) \,\, = \,\,  
\int_{\tilde 
M} \cI(\ell_i) - \cI(r_i)\, dM\, . 
\end{equation} 
\end{Corollary}
\par
\begin{Remark}
\label{RemII5.23}
\normalfont
The regions in both generalizations of Weyl\rq s formula and Steiner\rq s 
formula for different $i$ and $j$ will only intersect in lower dimensional 
regions.  Thus, in both cases we can sum the integrals on the RHS for 
multiple $i$ to obtain formulas for a union of $\gW_i$.
\end{Remark}
\par

\section{Positional Properties of Multi-Object Configurations}
\label{SecII:PosGeom}
\par 
In this section we define positional geometric invariants of 
configurations in terms of volumetric measures of associated regions 
defined by the linking structure.  We emphasize that the volumetric 
measures versus boundary measures of positional geometry have two 
advantages: 1) they are computable from the skeletal linking structure, 
and unlike surface measures, they do not require the partial Blum 
condition to compute the invariants; and 2) as in Figure \ref{fig.III1}, the 
volumetric invariants capture the total geometric structure of regions 
better than boundary measures.  \par   
We proceed as follows.  We first use the linking structure to determine 
which of the objects should be regarded as neighboring objects.  Then, we 
use the regions associated to the linking structure to define invariants 
which measure the closeness of such neighboring objects.  Second, we 
further introduce numerical invariants measuring the positional 
significance of objects for the configuration.  These allow us to identify 
which objects are central to the configuration and which ones are 
peripheral.  \par 
In \S \ref{SecII:TierGrph.PxoxMatr} we will use the closeness measures to 
construct a \lq\lq proximity matrix\rq\rq which yields proximity weights 
for the objects based on their closeness to other objects.  We also will 
use both types of invariants to construct a {\it tiered linking graph}, with 
vertices representing the objects, and edges between vertices 
representing neighboring objects, with the closeness and significance 
values assigned to the edges, resp. vertices.  By applying threshold values 
to this structure we can exhibit the subconfigurations within the given 
thresholds.  
\par

\subsection*{Neighboring Objects and Measures of Closeness} 
\par
We consider a configuration $\bgW = \{\gW_i\} \subset \tilde \gW$, with 
a bounded skeletal linking structure.  We use linking between objects 
$\gW_j$ and $\gW_i$ as a criterion for their being neighbors, so that 
objects which are not linked are not considered neighbors.  The simplest 
measure of closeness between neighboring objects is the minimum 
distance between the objects.  However, this ignores the size of the 
objects and how big a portion of each object is close to the other object, 
as illustrated in Figure~\ref{fig.II2} where $\gW_3$ is close to $\gW_1$ 
for a small region but $\gW_2$ is close to $\gW_1$ over a larger region.  
Moreover, if we choose a more global definition of closeness involving all 
neighboring boundary points, then as in Figure \ref{fig.III1}, this will not 
measure the portions of the objects which are close.  
We do overcome both of these issues by using volumetric measures of 
appropriate regions defined using the linking structures. 
\par
For a configuration $\bgW$ with a skeletal linking structure, we 
introduced in \S \ref{SII:sec.int} regions $\gW_{i \to j}$ and $\gW_{j \to 
i}$ which capture the neighbor relations between $\gW_i$ and $\gW_j$.  
Since $\cN_{i \to j}$ and $\cN_{j \to i}$ share a common boundary region 
in $M_0$, they are both empty if one is, and then both $\gW_{i \to j}$ and 
$\gW_{j \to i}$ are empty.  In that case $\gW_i$ and $\gW_j$ are not 
linked.  Otherwise, we may introduce a measure of closeness.  \par
\begin{figure}[!t]
\includegraphics[width=8cm]{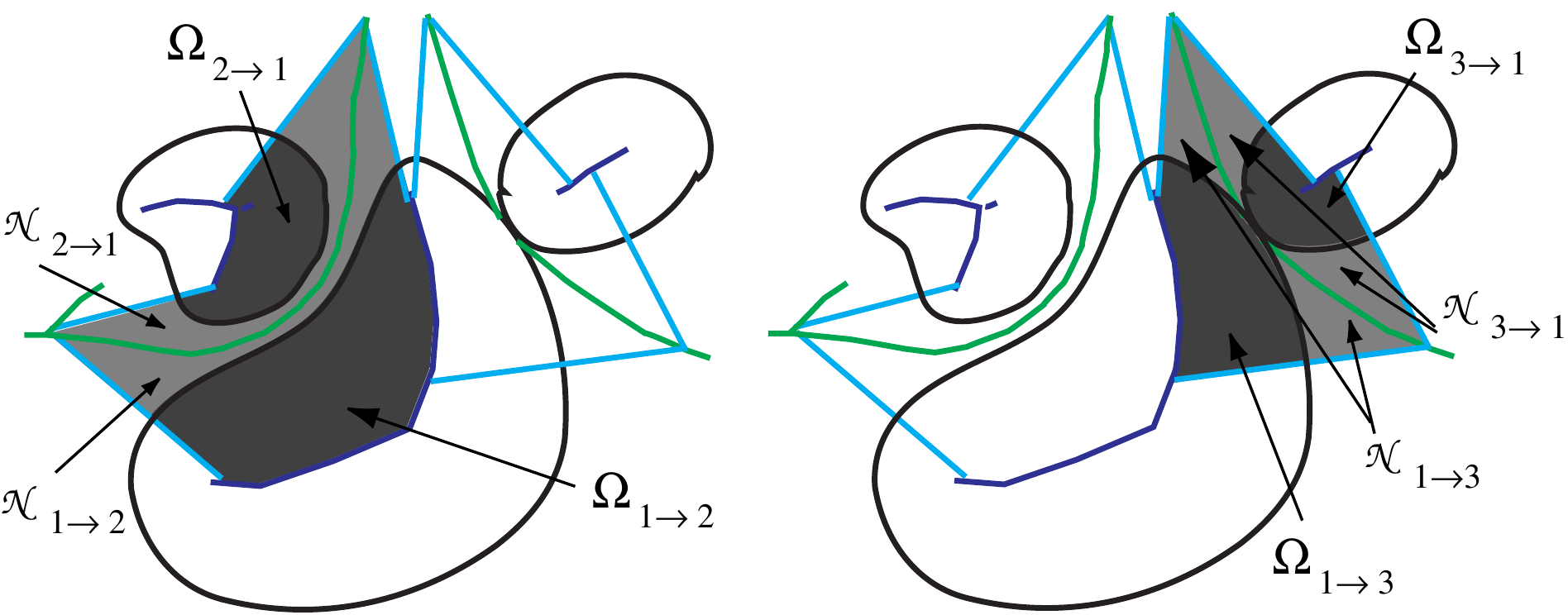}
\caption{\label{fig.II6.9} Measure of closeness for regions in the 
configuration in b) of Figure \ref{fig.II4.1} bounded via a threshold with 
the bounded skeletal linking structure.  For a pair of neighboring regions 
$\gW_i$ and $\gW_j$, $c_{i\to j}$ denotes the ratio of the volume of the 
darker region $vol(\gW_{i \to j})$ and the volume of the total shaded 
region $vol(\cR_{i \to j}) = vol(\gW_{i \to j}) + vol(\cN_{i \to j})$.}
\end{figure}

 There are two different ways to do this, each having a probabilistic 
interpretation.  First, we let 
$$ c_{i \to j} = \frac{\vol(\gW_{i \to j})}{\vol(\cR_{i \to j})}\qquad \text{ 
and } \qquad c_{i\, j} \, = \, c_{i \to j}\cdot c_{j \to i}\, .$$  
Then, $c_{i \to j}$ is the probability that a point chosen at random in 
$\cR_{i \to j}$ will lie in $\gW_i$ (see Figure~\ref{fig.II6.9}); so $c_{i\, 
j}$ is the probability that a pair of points, one each in $\cR_{i \to j}$ and 
$\cR_{j \to i}$ both lie in the corresponding regions $\gW_i$ and $\gW_j$.  
\par 
\par
\par
Note that $c_{i\, j}$ contains much more information than the closest 
distance between $\gW_i$ and $\gW_j$, and even the \lq\lq $L^1$-
measure\rq\rq of the region between $\gW_i$ and $\gW_j$.  It compares 
this measure with how much of the regions $\gW_i$ and $\gW_j$ are 
closest as neighbors.  If both $\gW_{i \to j}$ and $\gW_{j \to i}$ are 
empty, we let $c_{i \to j}$, $c_{j \to i}$, and  $c_{i\, j} = 0$.  Also, we let 
$c_{i\, i} = 1$.  Thus, from the collection of values $\{c_{i\, j}\}$ we can 
compare the closeness of any pair of regions.  
\par 
Since these invariants depend on a bounded skeletal linking structure, one 
way to introduce a parametrized family $c_{i\, j}(\tau)$ is by considering 
the varying threshold values $\tau$.  For example, $\tau$ may represent 
the maximum allowable values for $\ell_i$ or the maximum value of 
$\ell_i$ relative to some intrinsic geometric linear invariant of $\gW_i$.  
As $\tau$ increases, the bounded region increases and how $c_{i\, 
j}(\tau)$ varies indicates how the closeness of the regions varies when 
larger linking values are taken into account.  Thus, this provides a method 
to use the local properties of the skeletal linking structure to introduce a 
scale of local closeness.  \par
A second way to introduce a measure of closeness is to use an \lq\lq 
additive contribution\rq\rq\, from each region and define
$$ c_{i\, j}^a \,\, = \,\, \frac{\vol(\gW_{i \to j}) + \vol(\gW_{j \to 
i})}{\vol(\cR_{i \to j}) + \vol(\cR_{j \to i})} \, . $$
Here $c_{i\, j}^a$ is the probability that a point chosen in the region 
$\cR_{i \to j} \cup \cR_{j \to i}$ lies in the configuration, i.e. in $\gW_i 
\cup\gW_j$.  We also let $c_{i\, j}^a = 0$ if $\gW_i$ and $\gW_j$ are not 
linked; and we let $c_{i\, i}^a = 1$.  Again, to obtain a more precise 
measure of closeness, we can vary a measure of threshold $\tau$ and 
obtain a varying family $c_{i\, j}^a(\tau)$.  
The invariants satisfy $0 \leq c_{i\, j}, c_{i\, j}^a \leq 1$.  The value $0$ 
indicates no linking, for values near $0$, the regions are neighbors but 
distant so they are \lq\lq weakly linked,\rq\rq and for values close to 
$1$, the regions are close over a large boundary region and are \lq\lq 
strongly linked.\rq\rq  \par
There is a simple but crude relation between $c_{i\, j}^a$ and the pair 
$c_{i \to j}$ and $c_{j \to i}$:
$$  c_{i\, j}^a \,\, \leq  \,\, c_{i \to j} \, + \, c_{j \to i} \, . $$
As $c_{i\, j}^a \leq 1$, this is only useful when the two regions are 
weakly linked.  \par
\subsection*{Measuring Positional Significance of Objects Via Linking 
Structures} 

\par
\begin{figure}[!t]
\centering
\includegraphics[width=6.5cm]{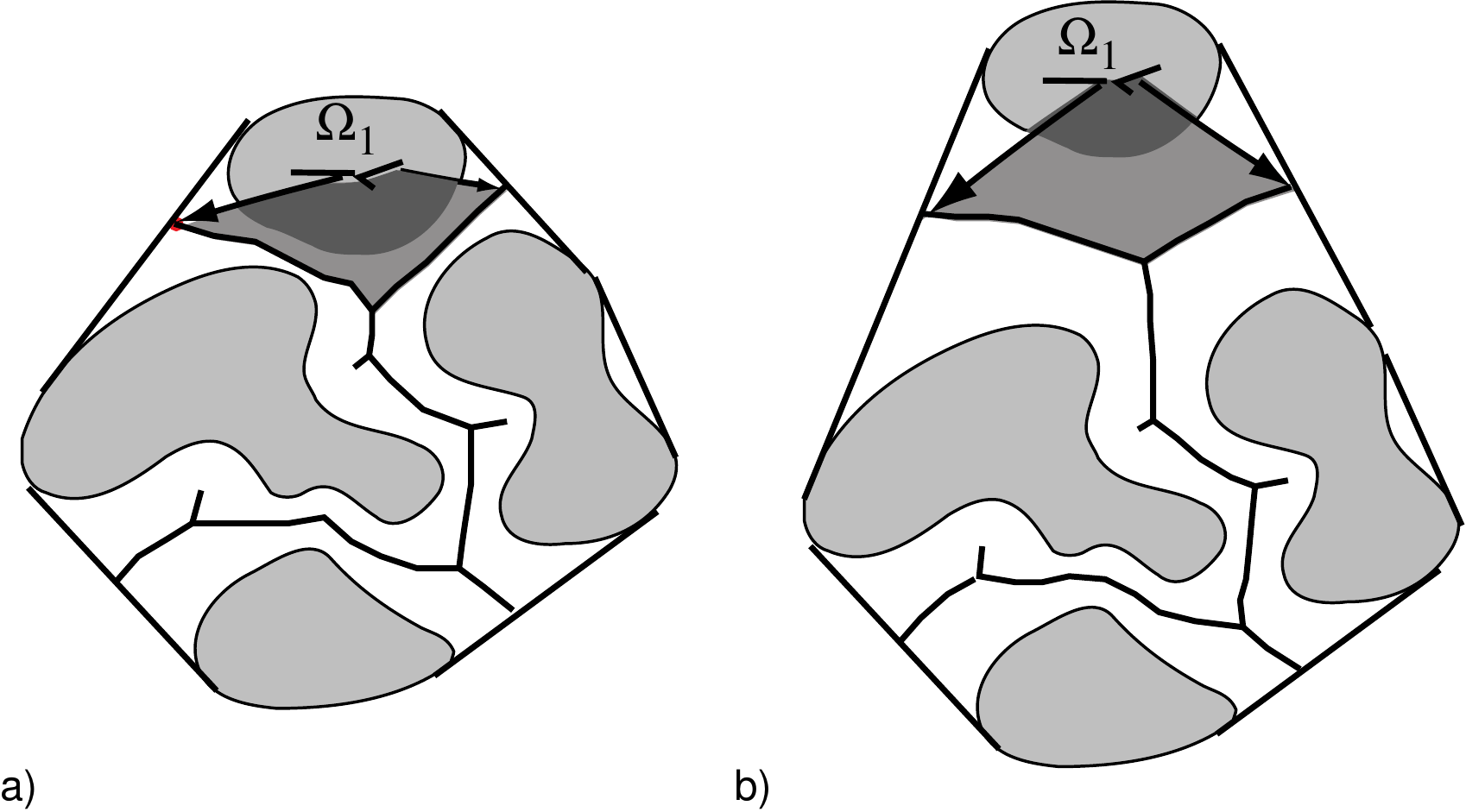}
\caption{\label{fig.II6.10} For $\gW_1$ in configurations using convex 
hull bounding regions, a measure of positional significance is the ratio of 
the darkest region to the union of the two more darkly shaded regions.  
In a) $\gW_1$ is central, while in b) when $\gW_1$ is moved away from 
the remaining regions, the ratio decreases indicating it is becoming more 
peripheral to the configuration.}
\end{figure}

In order to measure {\it positional significance} of an object among a 
collection of objects, we can think in both absolute and relative terms.  In 
each case, we emphasize that we are considering a form of geometric 
significance of objects relative to the configuration, rather than some 
other notion such as significance in the sense of statistics.  We begin 
with the relative version.  Given $\gW_i$, we define the {\it positional 
significance measure}
$$  s_i =  \frac{\sum_{j \neq i} \vol(\gW_{i \to j})}{\sum_{j \neq i} 
\vol(\cR_{i \to j})} \, . $$
It takes values $0 \leq s_i \leq 1$.  For values near $0$, the portion of the 
region of $\gW_i$ linked to other regions is a small fraction of the 
external region between $\gW_i$ and the other regions. Thus, compared to 
its size it is distant from other neighboring objects, so it is a peripheral 
region of the configuration.  We would have the value $s = 0$ if $\gW_i$ is 
not linked to any other region in $\tilde \gW$, which may occur if there is 
a threshold for which the region is not linked to another region with a 
linking vector of length less than the threshold.  By contrast, if $s_i$ is 
close to $1$, then there is very little external region between $\gW_i$ 
and the other regions.  Thus, $\gW_i$ is central for the configuration, 
see Figure~\ref{fig.II6.10}. 
Note that
$$  s_i \,\, \leq \,\, \sum_{j \neq i} c_{i\to j} \, , $$
so that $\gW_i$ being weakly linked to the other regions implies it has 
small significance for the configuration.  If we would like to further base 
the positional significance of the region $\gW_i$ on its absolute size, we 
can alternatively use an absolute measure of positional significance 
defined by $\tilde s_i = s_i \vol(\gW_i)$.  Then, the effect of the 
smallness of $s_i$ will be partially counterbalanced by the size of 
$\gW_i$.  
\par
\par

\subsection*{Properties of Invariants for Closeness and Positional 
Significance} 
\par
We consider three properties of these invariants: \par
\begin{itemize} 
\item[1)] computation of all of the invariants as skeletal linking 
integrals;
\item[2)] invariance under the action of the Euclidean group and scaling; 
and 
\item[3)] continuity of the invariants under small perturbations of generic 
configurations.
\end{itemize}
\par
\subsubsection*{\it Computation of the Invariants as Skeletal Linking 
Integrals}  
We can use the results from the previous section to compute as skeletal 
linking integrals the above volumes of regions associated to $\bgW$.  This 
is summarized by the following, see \cite[Thm. 11.3]{DG}.
\begin{Thm}
\label{ThmII6.3}
If $\bgW = \{\gW_i\} \subset \tilde \gW \subset \R^{n+1}$, for $n = 1$ or 
$2$ is a multi-object configuration, with a bounded skeletal linking 
structure, then all global invariants of the configuration which can be 
expressed as integrals over regions in 
$\R^2$, resp. $\R^3$, can be computed as skeletal linking integrals using 
Theorem \ref{ThmII5.5}.  In particular, the invariants $c_{i\to j}$, 
$c_{i\, j}$, $c^a_{i\, j}$, and $s_i$ are given as the quotients of two 
skeletal linking integrals using (\ref{EqnII5.20}) and (\ref{EqnII5.20b}). 
\end{Thm}  
\par 
\begin{Remark}
\label{RemII5.2} 
\normalfont We emphasize that we could try to alternatively use 
boundary measures for the regions to define closeness and significance.  
There are two problems with this approach.  From a computational point of 
view, the skeletal structures could only be used where the partial Blum 
condition is 
satisfied.  Moreover, boundary measures do not capture how much of the 
regions are close to each other (only where their boundaries are close).  
For these reasons we have concentrated on (ratios of) volumetric 
measures to capture positional geometry of the configuration.   
\end{Remark}
\par
\subsubsection*{\it Invariance Under the Action of the Euclidean Group 
and 
Scaling}  \par
Second, we establish the invariance of the invariants defining closeness 
and significance under Euclidean motions and scaling.  Let $\bgW = 
\{\gW_i\} \subset \tilde \gW \subset \R^{n+1}$, for $n = 1$ or $2$ be a 
multi-object configuration, with a 
bounded skeletal linking structure $\{(M_i, U_i, \ell_i)\}$.  If $f$ is a 
Euclidean 
motion and $a > 0$ is a scaling factor, then we may let 
$\bgW^{\prime} = \{\gW_i^{\prime}\} \subset \tilde \gW^{\prime}$, where 
$\gW_i^{\prime} = f(\gW_i)$ and $\tilde \gW^{\prime} = f(\tilde \gW)$.  
We also let $\{(M_i^{\prime}, U_i^{\prime}, \ell_i^{\prime})\}$ be a 
skeletal linking structure for $\bgW^{\prime}$ defined by $M_i^{\prime} = 
f(M_i)$, $U_i^{\prime} = f(U_i)$, and $\ell_i^{\prime} = \ell_i$.  As $f$ 
preserves distance and angles, we have $r_i^{\prime} = r_i$, and the image 
of the linking flow for $\bgW$ is the linking flow for $\bgW^{\prime}$. 
Then, $\{(M_i^{\prime}, U_i^{\prime}, \ell_i^{\prime})\}$ satisfies the 
conditions for being a skeletal linking structure for $\bgW^{\prime}$.  As 
$\tilde \gW^{\prime} = f(\tilde \gW)$, the corresponding bounded linking 
structure for $\bgW^{\prime}$ using $\tilde \gW^{\prime}$ is the image of 
that for $\bgW$ for $\tilde \gW$.   Then, the associated linking regions 
for $\bgW^{\prime}$ are the images of the corresponding associated 
linking regons for $\bgW$.  Since $f$ preserves volumes, the invariants 
for closeness and significance are preserved by $f$.  \par 
If instead we consider a scaling by the factor $a > 0$, then we let $g_a(x) 
= a\cdot x$.  Now the images of $\bgW$ and $\tilde \gW$ under $g_a$ 
define a configuration $\bgW^{\prime}$ in $\tilde \gW^{\prime}$.  We 
likewise let $\{(M_i^{\prime}, U_i^{\prime}, \ell_i^{\prime})\}$ be defined 
by $M_i^{\prime} = g_a(M_i)$, $U_i^{\prime} = a U_i$, and $\ell_i^{\prime} 
= a \ell_i$ (and $r_i^{\prime} =a r_i$).  As before this is a skeletal 
structure for $\bgW^{\prime}$.  Everything goes through except that $g_a$ 
multiplies volume by $a^{n+1},n=1,2$.  However, as the invariants are 
ratios of 
volumes, they again do not change.  We summarize this with the following.
\begin{Proposition}
\label{PropII6.4}
If $\bgW = \{\gW_i\} \subset \tilde \gW \subset \R^{n+1}$, for $n = 1$ or 
$2$ is a multi-object configuration, with a skeletal linking structure, 
then the invariants $c_{i\to j}$, $c_{i\, j}$, $c^a_{i\, j}$, and $s_i$ are 
invariant under the action of Euclidean 
motion and scaling applied to both $\bgW$ and $\tilde \gW$ for the image 
of the skeletal linking structure for the image configuration and bounding 
region. 
\end{Proposition}
\par 
We note that if we consider the absolute significance $\tilde s_i$, then it 
is still invariant under Euclidean motions. However, under scaling by $a > 
0$, it changes by the factor $a^{n+1}$ for $n=1,2$; but this would not alter 
the 
hierarchy based on absolute significance, as all $\tilde s_i$ would be 
multiplied by the same factor.
\begin{Remark}
\label{RemII.6.5}
\normalfont
Importantly, the invariance in Proposition \ref{PropII6.4} crucially 
depends on also applying the Euclidean motion and/or scaling to the 
bounding region $\tilde \gW$.  For either thresholds or convex hulls, there 
is no problem.  If instead the bounding region is either fixed, or depends 
upon an external condition which prevents it from transforming 
along with the configuration, then the invariance does not hold.  The 
change under Euclidean motion $f$ or scaling by $a$ could be measured in 
terms of the changes in the portions of the linking regions that lie in the 
difference region between $\tilde \gW$ and its image under $f$ or scaling.
\end{Remark}
\par  
\subsubsection*{\it Continuity and Changes Under Small Perturbations} 
\par
Lastly, suppose that the configuration $\bgW = \{\gW_i\} \subset \tilde 
\gW$ has a skeletal linking structure.  We ask how  will the invariants 
change under small perturbations?   \par
First, if objects undergo a sufficiently small deformation, then we may 
deform the skeletal linking structure to be the skeletal linking structure 
for the deformed configuration, in such a way that the skeletal structures 
and linking vector fields will deform in a stratawise differentiable 
fashion.  Then, the associated regions will also deform in a piecewise 
differentiable fashion.  Hence, the volumes of these regions will vary 
continuously.  Thus, the quotients of the volumes will also vary 
continuously.  It then follows that the invariants, which are quotients of 
such volumes will also vary continuously. 
\par
How exactly they will vary will depend on the particular deformation.  For 
example, suppose we enlarge one of the regions $\gW_i$ by increasing the 
radial vectors by a factor $a > 1$, so that $a r_i < \ell_i$, and without 
altering the remainder of the skeletal structure.  If the region remains in 
the bounding region and doesn\rq t intersect itself or other regions, then 
the ratio $\vol(\gW_{i \to j})$ to $\vol(\cR_{i \to j})$ will increase for 
each $j$ so the $s_i$ will increase, as will the $c_{i\to j}$.  If instead $0 
< a < 1$, then $s_i$ and $c_{i\to j}$ will decrease.  If we move the 
region $\gW_i$ in a direction away from all of the 
other regions without altering its size, then in general $s_i$ will 
decrease, and conversely if we move it toward the other regions, generally 
$s_i$ will increase. Thus, the changes in the invariants capture the 
changes in the configuration resulting from the deformation.

\section{Proximity Matrix and Tiered Linking Graph for Multi-Object 
Configurations}
\label{SecII:TierGrph.PxoxMatr}
The invariants we introduced in the previous section individually capture 
positional properties of objects in a configuration.  We show how they 
taken together provide numerical structures which summarize the 
relations in the configuration.  These take two forms: a {\em proximity 
matrix} that has a unique postive eigenvalue with eigenvector with 
positive entries assigning unique {\em proximity weights} to the objects 
based on their relative closeness; and a {\em tiered linking graph} 
structure, which identifies substructures satisfying threshold conditions.  
\subsection*{Proximity Matrix and Proximity Weights}
\par
We consider a configuration $\bgW = \{ \gW_i\}$ with a bounded skeletal 
linking structure.  If there are $n$ objects,  we let $P$ be the $n \times n$ 
matrix with entries $c_{i\, j}$ given in the previous section (or instead 
with $c^a_{i\, j}$).  We refer to $P$ as the {\em proximity matrix}.  The 
proximity matrix purely measures the relative amounts of two objects 
which are neighbors compared with their adjacent regions. We can further 
weight the proximity matrix to take into account the size of the objects.  
If for each $i$, $V_i > 0$ is a measure of the object $\gW_i$, and we let 
$V_{tot} = \sum_{i = 1}^{n} V_i$, then for $v_i = \frac{V_i}{V_{tot}}$, the 
vector $\bv = (v_1, v_2, \dots , v_n)$ is a positive weight vector for the 
relative portion each $\gW_i$ contributes to the total measure $V_{tot}$.  
Two possibilities for $V_i$ are either the total volume/area 
$\vol(\gW_i)$, yielding the vector $\bv_{vol}$ or instead $V_i = \sum_{j 
\neq i}\vol(\gW_{i \to j})$, the portion of $\gW_i$ which is linked to 
some other object in the configuration, yielding $\bv_{lk}$.  Then, we can 
form the { \em renormalized proximity matrix} $\tilde P = (c_{i\, 
j}\frac{v_i}{v_j})$.  The proximity and renormalized proximity matrices 
yield {\em proximity weights} and {\em renormalized proximity weights} 
for the objects in the configuration as follows.

\begin{Proposition}
\label{Prop6.1}
Consider a configuration of objects $\bgW = \{ \gW_i\}$ with a bounded 
skeletal linking structure, such that within a bounding region there is no 
subset of objects all of which are unlinked to the complementary set of 
objects in the configuration.  Then, 
\begin{itemize}
\item[i)] both the proximity matrix and renormalized matrix have the 
same unique maximal positive eigenvalue $\gl_P$ with an eigenvector 
$\bw$ for $P$ and $\tilde{\bw}$ for $\tilde P$ both having all positive 
entries; 
\item[ii)]  this is the only eigenvalue for either matrix with an 
eigenvector with these properties; and 
\item[iii)] if $\bw = (w_1, \dots , w_n)$ is such an eigenvector for $P$, 
then $\tilde{\bw} = (w_1 V_1, \dots , w_n V_n)$ is such an eigenvector 
for $\tilde P$.
\end{itemize}
\end{Proposition}
\par 
Since $\bw$ and $\tilde{\bw}$ are only well-defined up to positive scalar 
multiples, we may normalize each to vectors $\bw_P = (w_1, \dots , 
w_n)$, resp. $\tilde \bw_P = (\tilde w_1, \dots , \tilde w_n)$ with 
$\sum_{i = 1}^{n} w_i = 1$, resp. $\sum_{i = 1}^{n} \tilde w_i = 1$.  Thus, 
each vector uniquely assigns a weight $w_i$, resp. $\tilde w_i$, to the 
object $\gW_i$ depending on the proximity of the other regions to it.  We 
refer to these weights as the {\em proximity weights}, resp. {\em 
renormalized proximity weights} for the objects in the configuration.  The 
proximity weights uniquely provide an ordering on the objects based on 
their proximity to other objects, with the renormalized weights modifying 
this weighting to include a measure of the size of the objects. 
\begin{proof}
By the properties of the closeness measures $c_{i\, j}$, the matrix $P$ 
has the properties that it is a symmetric nonnegative matrix.  Moreover, 
because of the properties of the configuration in the bounding region, the 
matrix $P$ is \lq\lq irreducible\rq\rq, which for nonnegative symmetric 
matrices reduces to the condition that for any $i$ there is a  $j \neq i$, 
such that $c_{i\, j} \neq 0$.  Then we may apply a version of the 
Perron-Frobenius theorem for irreducible nonnegative matrices, see e.g. 
\cite{Gm} or \cite[Chap. 8]{Me}, to conclude there is a unique largest 
positive eigenvalue $\gl_P$ for $P$ with eigenvector $\bw$ with all 
positive entries.  Moreover, this is the only eigenvalue with an eigenvector 
with these properties.  \par 
As $P$ is irreducible, so is $\tilde P$, which is conjugate to $P$ by a 
diagonal matrix with values $\frac{1}{v_i}$ on the diagonal.  Furthermore 
it follows that the eigenvalues of $\tilde P$ are the same as those of $P$, 
and for a common eigenvalue $\gl$, with the eigenvector $\bw = (w_1, 
\dots , w_n)$ for $P$, there is a corresponding eigenvector $\tilde \bw = 
(w_1 V_1, \dots , w_n V_n)$ for $\tilde P$.  Thus, the results also follow 
for $\tilde P$.
\end{proof}
\par
\begin{figure}[!t]
\begin{center} 
\includegraphics[width=4.25cm]{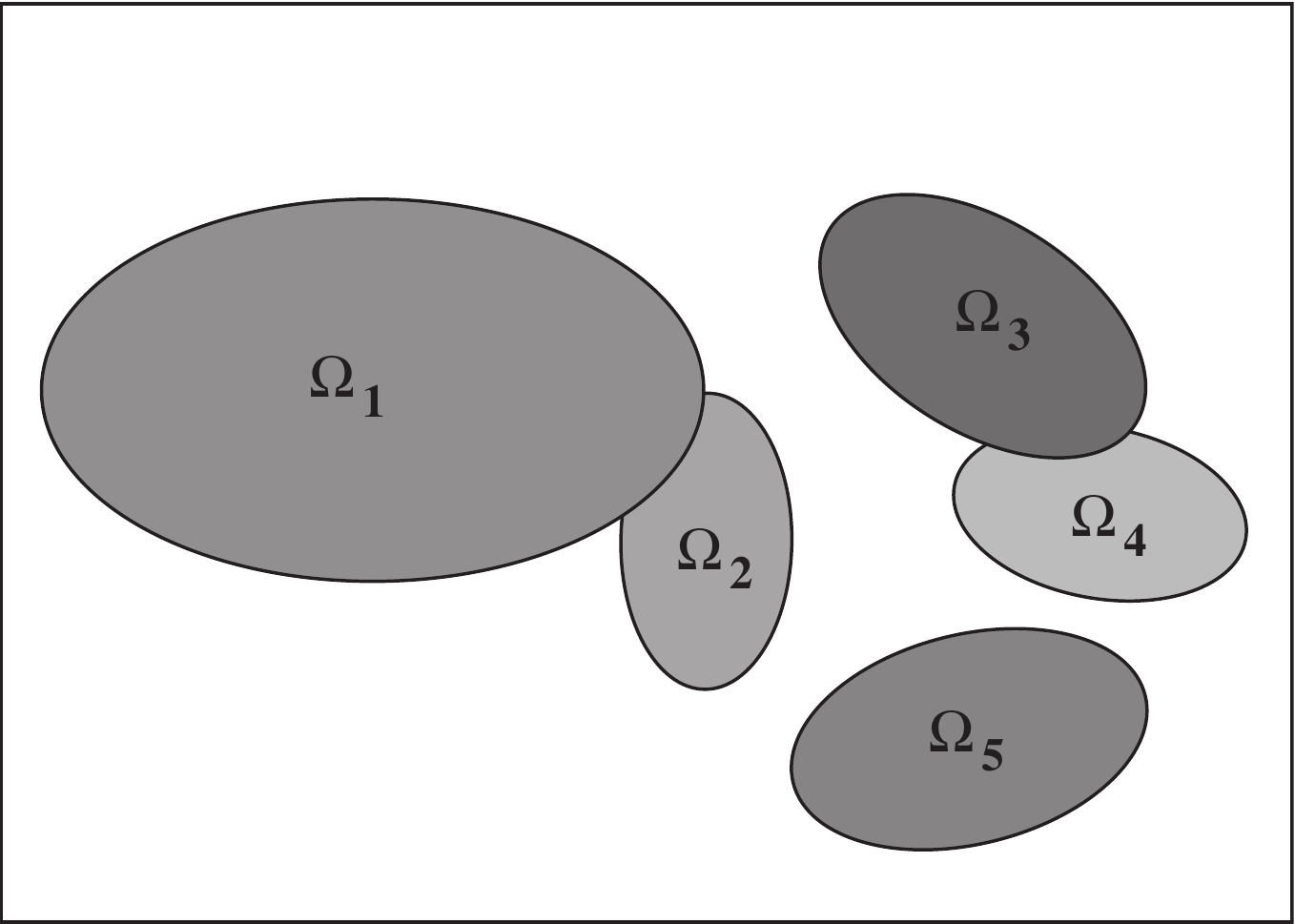}
\hspace*{0.05cm}
\includegraphics[width=4.25cm]{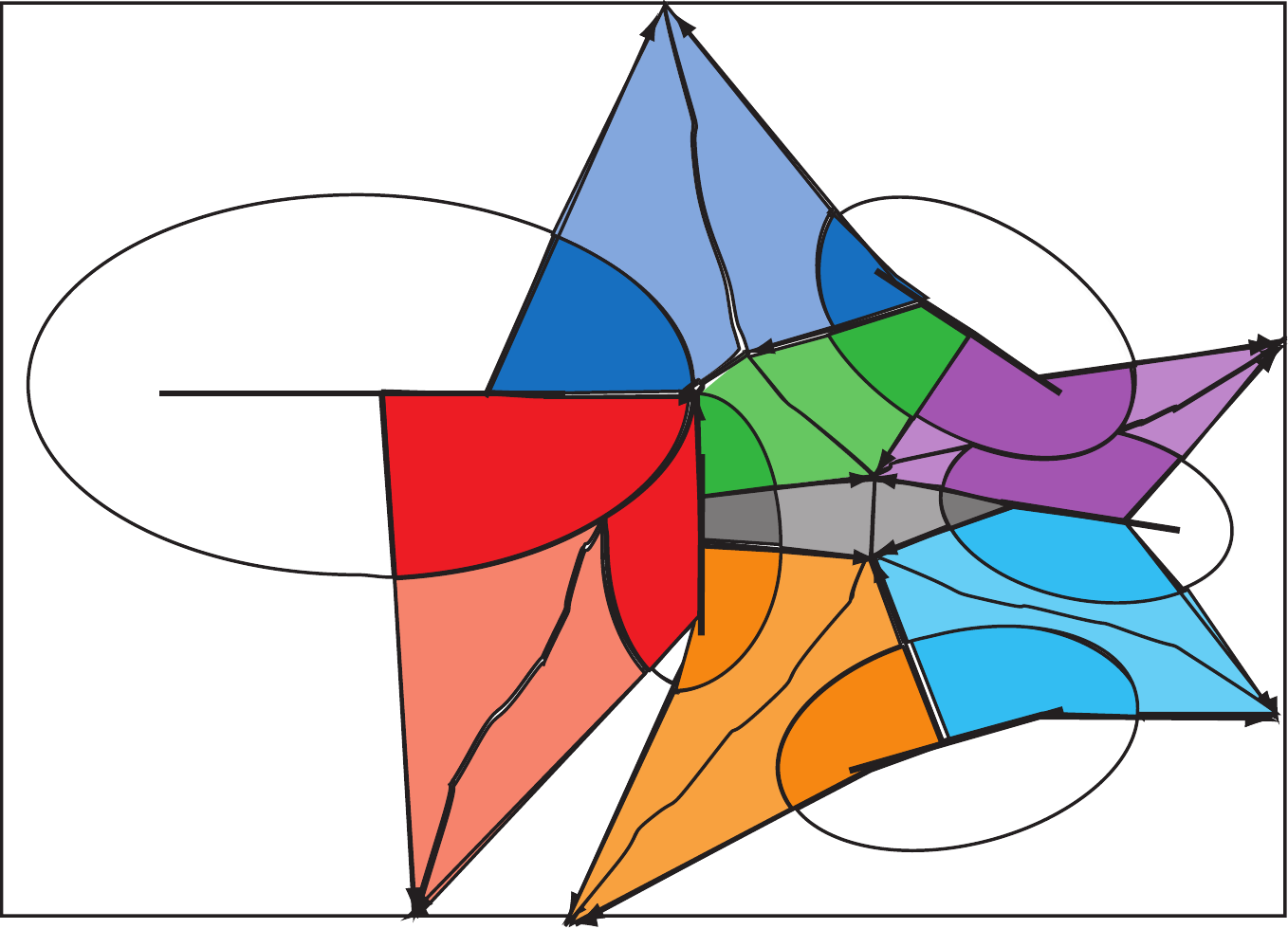}
\end{center}
a) \hspace{1.5in} b) \hspace {1.5in}
\caption{\label{fig.II8} In a) is a synthetic configuration of 5 objects in a 
bounding window. In b) is the decomposition of the neighboring regions 
based on a  Blum linking structure which is deformed at the singular 
points of the boundaries.  The subregions $\gW_{i\to j}$ and $\gW_{j\to 
i}$ of different objects $\gW_{i}$ and $\gW_{j}$ which are linked have the 
same color, and the external linking regions $\cN_{i\to j}$ and $\cN_{j\to 
i}$ have the same lighter color.  The white regions of objects are unlinked 
in the bounding window.  The values for closeness, significance, the 
proximity matrix, the proximity weights and the renormalized proximity 
weights in Example \ref{Ex6.1} are computed from this figure.}  
\end{figure}
\par

\begin{Example}
\label{Ex6.1}
\normalfont In a) of Figure \ref{fig.II8} is a synthetic configuration of $5$ objects in a 
bounding window in $\R^2$.  In this case \lq\lq vol\rq\rq\, refers to the 
area of the various regions.  The positional invariants are computed using 
the linking regions and neighborhoods indicated in b) of Figure 
\ref{fig.II8}. This yields: the proximity matrix $P$ given in (\ref{Eqn6.2}) 
and in (\ref{Eqn6.3}) the normalized measures for either the areas of the 
individual object regions, $\bv_{vol}$, or the areas of the total internal 
linking regions for each object, $\bv_{lk}$, and the positional significance 
vector $\bs$ consisting of the positional significance value for each 
object. \par

\begin{equation}
\label{Eqn6.2}
P \,\, = \,\, 
\begin{pmatrix} 
1        &   .240    &  .081    &     0     &    0     \\  
.240   &       1     &  .200    &  .104   &   .152  \\
.081   &   .200    &     1      &  .487   &    0      \\
0        &    .104   &   .487   &    1      &   .305   \\
0        &    .152   &     0      &   .305  &      1 
 \end{pmatrix}
\end{equation}
The proximity weight vector given by Proposition \ref{Prop6.1} is 
computed to be 
\begin{equation}
\label{Eqn6.2a}  
\bw_P \,\, = \,\, (0.10,  0.19,  0.26,  0.28,  0.17) \, .  
\end{equation}
\par 
We first note that despite $\gW_1$ having by far the largest area, its 
weight when determined from pure closeness data is small compared to 
the other objects.  This is because as we see in b) of Figure \ref{fig.II8}, 
most of $\gW_1$ is unlinked and hence effectively invisible to the other 
objects.  By contrast, a much greater portion of $\gW_3$ and $\gW_4$ 
plays a central role in the configuration.
We compare these weights with the renormalized weights using the 
weight vectors $\bv_{vol}$ and $\bv_{lk}$ given by (\ref{Eqn6.3}).
\begin{align}
\label{Eqn6.3}
\bv_{vol} \,\, &= \,\, (0.49,  0.11,  0.15,  0.10,  0.15) \, , \notag \\
\bv_{lk} \,\, &= \,\,  (0.34,  0.21,  0.10,  0.16,  0.19)  \, , \\
\bs \,\, &= \,\,  (0.52,  0.40,  0.43,  0.46,  0.49) \, .  \notag 
\end{align}
Using the weight vectors $\bv_{vol}$ and $\bv_{lk}$ given by  
(\ref{Eqn6.3}), we obtain the renormalized weight vectors for the 
configuration in (\ref{Eqn6.4}) 
\begin{align}
\label{Eqn6.4}
\tilde \bw_{P, \, vol} \,\, &= \,\, (0.30,  0.12,  0.24,  0.19,  0.15) \, ,   
\notag \\
\tilde \bw_{P, \,lk} \,\, &= \,\,  (0.18,  0.22,  0.15,  0.27,  0.18)   \, . 
\end{align}
\par
We now see that using the vector $\bv_{lk}$, using areas of the linked 
regions of the objects, the weight of $\gW_1$ increases and $\gW_3$ 
significantly decreases, while the others change only slightly.  If instead 
we renormalize by the total areas of the objects given by $\bv_{vol}$, 
then the overall importance of $\gW_1$, as measured by its size becomes 
evident in $\tilde \bw_P$.  Thus, the three vectors $\bw_P$, $\tilde 
\bw_{P, \, vol}$, and $\tilde \bw_{P, \, lk}$ give different measures of 
the weights of the objects within the configuration, successively 
increasing the importance attached to the size of the objects involved in 
the configuration.  We may compare these three measures with the 
positional significance measure given in $\bs$ in (\ref{Eqn6.3}), where 
despite the differences in size, number of neighbors, and linking 
structures, the calculated significance measures for all five objects are 
within a narrow range $0.4 \leq s_i \leq 0.52$.  \par
Viewing b) of Figure \ref{fig.II8}, we remark that alternative methods 
that would reduce the effect of the triangular regions reaching out to the 
boundary would be either to use a threshold for the linking functions 
$\ell_i$ or use the convex hull of the configuration as the bounding region 
as described in \cite[\S 3]{DG1}.
\end{Example}

\subsection*{Tiered Linking Graph}
\par
We can furthermore use the invariants $c_{i\, j}$ and $s_i$ in a second 
way to construct a {\em tiered graph} which simultaneously captures both 
the relative positions of the objects and their significance for the 
configuration.  For us a graph $\gG$ is defined by a finite set of vertices 
$V = \{v_i: i = 1, \dots , m\}$, and a set of unordered edges $E = \{e_{i\, 
j}\}$ with at most one edge $e_{i\, j}$ between any pair of distinct 
vertices $v_i$ and $v_j$. 
\begin{Definition}
A {\em tiered graph} consists of a graph $\gG$ together with a discrete 
nonnegative function $f: V \cup E \to \R_{+}$ which we shall more simply 
denote by $f: \gG \to \R_{+}$.  The discrete function $f$ has values $f(v_i) 
= a_i \geq 0$ for each vertex $v_i$, and $f(e_{i\, j}) = b_{i\, j} \geq 0$ for 
each edge $e_{i\, j}$.  
\end{Definition}
\par
Given such a tiered graph, we can view its values on vertices and edges as 
height functions assigning weights to the vertices and edges; and then 
apply \lq\lq thresholds\rq\rq to $f$ to identify subgraphs, distinguished 
vertices and edges.  First, given a value $b > 0$, we can consider the 
subgraph $\gG_b$ consisting of all vertices, but only those edges where $f 
\geq b$.  $\gG_b$ decomposes into connected subgraphs consisting of 
vertices which have edges of weights $> b$.  As $b$ decreases from $B = 
\max \{b_{i\, j}\}$, then we see the smaller graphs begin to merge as 
edges are added, until we reach $\gG$ for $b = \min \{b_{i\, j}\}$.  \par 
If instead we consider the threshold $a$ for $f$ on vertices, then instead 
we define $\gG^a$ to consist of those vertices with $f \geq a$, and only 
those edges joining two vertices within this set.  This identifies a 
subgraph consisting of the most important vertices as measured by 
weights, along with the edges between these vertices.  Again as $a$ 
decreases from $A = \max \{a_i\}$, then again we see the small graphs 
being supplemented by additional vertices with edges being added from 
these vertices until we reach the full graph when $a = \min \{a_i\}$.  This 
gives a hierarchical structure to the graph $\gG$.  Along with the 
subgraphs and the hierarchical structure, we can also identify vertices 
which are joined by strongly weighted edges, and important vertices with 
large weights $a_i$, and less significant ones with small weights $a_i$. 
\par
\begin{figure}[!t]
\centering
\includegraphics[width=4.5cm]{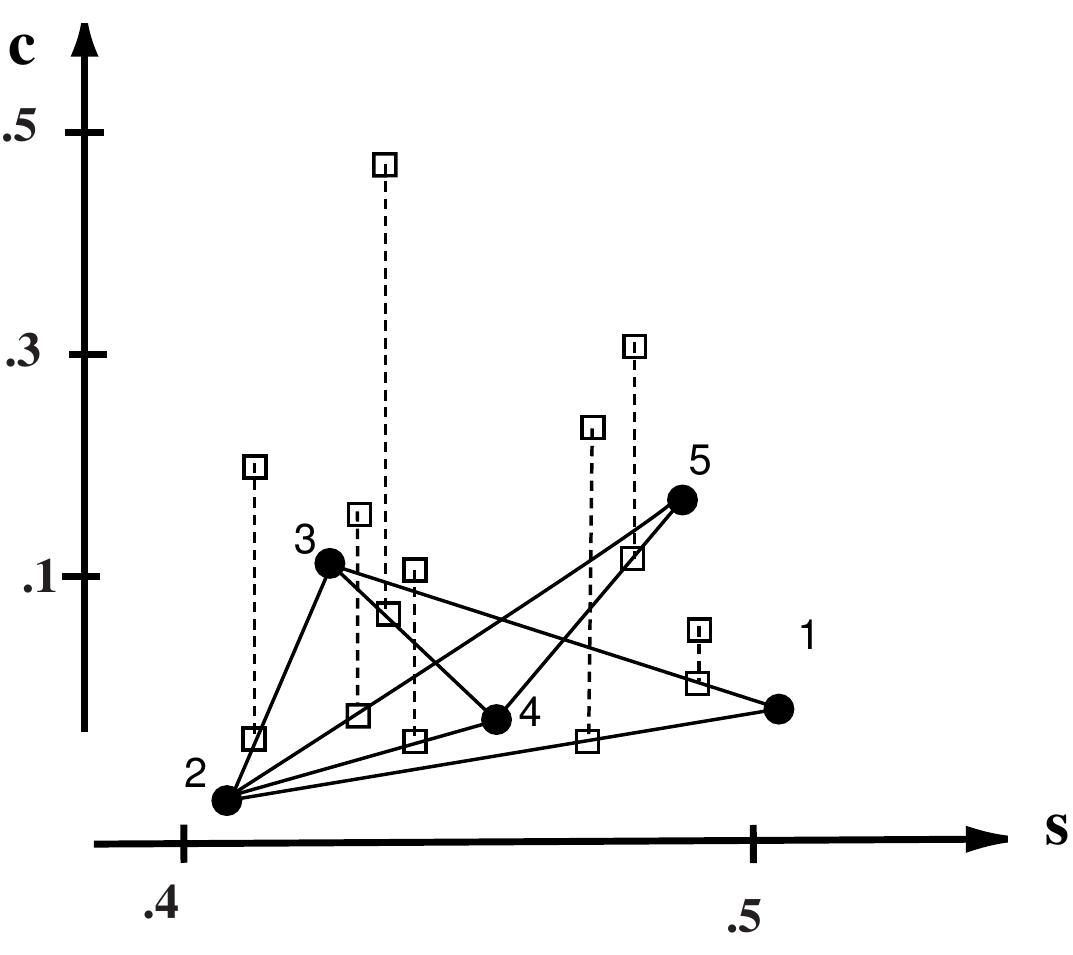}
\caption{Example of the tiered graph structure for the configuration in 
Example \ref{Ex6.1} shown in Figure \ref{fig.II8}. The horizontal axis 
indicates positional significance $s$, and closeness $c_{i\, j}$ is given by 
the values of the height function on the edges indicated by the tops of the 
dotted lines.}
\label{fig.II6.11}
\end{figure}
\par
This approach, using the tiered graph structure, applies to a configuration 
of multiple objects with a skeletal linking structure.  We define the 
associated {\it tiered linking graph} $\gL$ as follows.  For each object 
$\gW_i$, we assign a vertex $v_i$ in $\gL$, and to each pair of 
neighboring objects $\gW_i$ and $\gW_j$, we assign an edge $e_{i\, j}$ 
joining the corresponding vertices.  If the objects are not neighbors, there 
is no edge.  We define the height function $f$ by: $f(v_i) = s_i$ and 
$f(e_{i\, j}) = c_{i\, j}$ (or $c^a_{i\, j}$).  
\par
An example is shown in Figure~\ref{fig.II6.11} for the configuration in 
Example \ref{Ex6.1} using the closeness and positional significance 
measures computed in the proximity matrix $P$ and positional 
significance vector $\bs$.  When we apply the thresholds, we remove 
vertices to the left of some vertical line or edges whose heights are 
below some height.  We see how subconfigurations associated to the 
subgraphs merge into larger configurations as the vertical line indicating 
$s$ moves to the right adding objects, or the height moves downwards, 
adding edges, with the resulting graphs being based on closeness or 
significance of the subconfigurations of objects.  Position along the 
$s$-axis identifies the hierarchy of objects in the configuration.  \par
For Example \ref{Ex6.1}, in Figure~\ref{fig.II6.11} we see that while  
$\gW_1$ has the greatest positional significance measure, the combined 
position and size of $\gW_5$ places it second.  Also, as we move upward 
along the closeness scale we remove edges; so for example, as we move 
above the closeness threshold of $0.2$, we see the subconfigurations of 
$\{ \gW_1, \gW_2\}$ and $\{ \gW_3, \gW_4, \gW_5\}$ appearing. 
\par
\subsection*{Concluding Remarks}
\par
Presently, the investigation of configurations of objects in images 
typically involves many ad hoc choices.  To approach such collections in a 
systematic way, there is needed a uniform approach based on structures 
whose properties allow investigators to associate numerical measures 
which capture geometric features of the configuration and which can then 
be compared for statistical purposes for various image processing goals.  
In this paper we have made use of a medial/skeletal linking structure to 
model such a configuration, as introduced in \cite{DG1}.  Using this 
structure we introduced a number of numerical invariants which capture 
positional geometry of the configuration, along with the geometric 
properties of the individual objects in the configuration.  These yield a 
collection of mathematical tools that have already been successfully 
applied to single objects in medical images and now have a rigorous 
mathematical form for being applied to entire configurations of objects.

\end{document}